\documentclass[10pt,journal,compsoc]{IEEEtran}
\usepackage{amsmath,amsfonts}
\usepackage{bbm}
\usepackage{algorithmic}
\usepackage{array}
\usepackage{textcomp}
\usepackage{stfloats} 
\usepackage{subfigure}
\usepackage{url}
\usepackage{verbatim}
\usepackage{color}
\usepackage{graphicx}
\usepackage[lined,boxed,commentsnumbered,ruled,linesnumbered]{algorithm2e} 
\newtheorem{theorem}{Theorem}
\newtheorem{myDef}{Definition}
\def\BibTeX{{\rm B\kern-.05em{\sc i\kern-.025em b}\kern-.08em
    T\kern-.1667em\lower.7ex\hbox{E}\kern-.125emX}}
\usepackage{balance}
\begin{document}
\title{Orchestrating Joint Offloading and Scheduling for Low-Latency Edge SLAM}

\author{
\IEEEauthorblockN{Yao Zhang, 
Yuyi Mao,  \IEEEmembership{Senior Member, IEEE},
Hui Wang,
Zhiwen Yu, \IEEEmembership{Senior Member, IEEE},
Song Guo, \IEEEmembership{Fellow, IEEE},
Jun Zhang,  \IEEEmembership{Fellow, IEEE},
Liang Wang,
Bin Guo,  \IEEEmembership{Senior Member, IEEE}}
\thanks{This work was supported in part by the National Key $R\&D$ Program of China (No. 2024YFB4505502), in part by the National Science Fund for Distinguished Young Scholars (62025205), in part
by the National Natural Science Foundation of China  (No. 62302396, 62332014), in part by the Natural Science Foundation of Shaanxi Province for Distinguished Young Scholars (No. 2023-JC-JQ-54),
in part by the fundings from the Hong Kong RGC General Research Fund (152244/21E, 152169/22E, 152228/23E, 162161/24E), Research Impact Fund (No. R5009-21, No. R5011-23F, No. R5060-19), Collaborative Research Fund (No. C1042-23GF), NSFC/RGC Collaborative Research Scheme (No. CRS\_HKUST602/24), Theme-based Research Scheme (No. T43-518/24-N), Areas of Excellence Scheme (No. AoE/E-601/22-R), and the InnoHK (HKGAI). (Corresponding author: Zhiwen Yu Email: zhiwenyu@nwpu.edu.cn)
}
\thanks{Y. Zhang, H. Wang, L. Wang, and B. Guo are with School of Computer Science, Northwestern Polytechnical University, Xi'an, Shaanxi, China.}%
\thanks{Y. Mao is with the School of Computer Science and Engineering, Macau University of Science and Technology, Taipa, Macau.}
\thanks{Z. Yu is with Harbin Engineering University, Harbin, Heilongjiang, China,
and also with the School of Computer Science, Northwestern Polytechnical
University, Xi'an, Shaanxi, China.}
\thanks{S. Guo, J. Zhang are with The Hong Kong University of Science and Technology, Hong Kong, China.}
}


\maketitle

\begin{abstract}
Visual Simultaneous Localization and Mapping (vSLAM) is a prevailing technology for many emerging robotic applications. Achieving real-time SLAM on mobile robotic systems with limited computational resources is challenging because the complexity of SLAM algorithms increases over time. This restriction can be lifted by offloading computations to edge servers, forming the emerging paradigm of \emph{edge-assisted SLAM}. 
Nevertheless, the exogenous and stochastic input processes affect the dynamics of the edge-assisted SLAM system. {Moreover, the requirements of clients on SLAM metrics change over time, exerting implicit and time-varying effects on the system.} In this paper, we aim to push the limit beyond existing edge-assist SLAM by proposing a new architecture that can handle the input-driven processes and {also satisfy clients' implicit and time-varying requirements.}  
The key innovations of our work involve a regional feature prediction method for importance-aware local data processing, a configuration adaptation policy that integrates data compression/decompression and task offloading, and an input-dependent learning framework for task scheduling with constraint satisfaction. Extensive experiments prove that our architecture improves pose estimation accuracy and saves up to $47\%$ of communication costs compared with {a popular edge-assisted SLAM system, as well as effectively satisfies the clients' requirements.}

\end{abstract}


\section{Introduction}
The recent advancements in sensing, communication, control, and Artificial Intelligence (AI), are accelerating the ubiquitous and human-centered applications of intelligent robotic systems, such as smart manipulators, surgical robots, autonomous vehicles, and drones \cite{kim2020research,liu2024crowdtransfer}. {Those applications usually execute computation-intensive algorithms in resource-limited devices or mobility scenarios.} SLAM is a popular vision algorithm that can provide accurate localization information and a spatial map of the environment by processing vision data, such as video frames captured by onboard cameras \cite{durrant2006simultaneous}. Due to its stable performance in facilitating seamless indoor-outdoor navigation and providing centimeter-level localization accuracy \cite{taketomi2017visual}, SLAM has become an indispensable technology that supports numerous emerging robotic applications, such as mobile augmented reality \cite{braud2020multipath} and high-definition (HD) map \cite{ahmad2020carmap}. 

\begin{figure}[!t]
\centering
\includegraphics[width=1\linewidth]{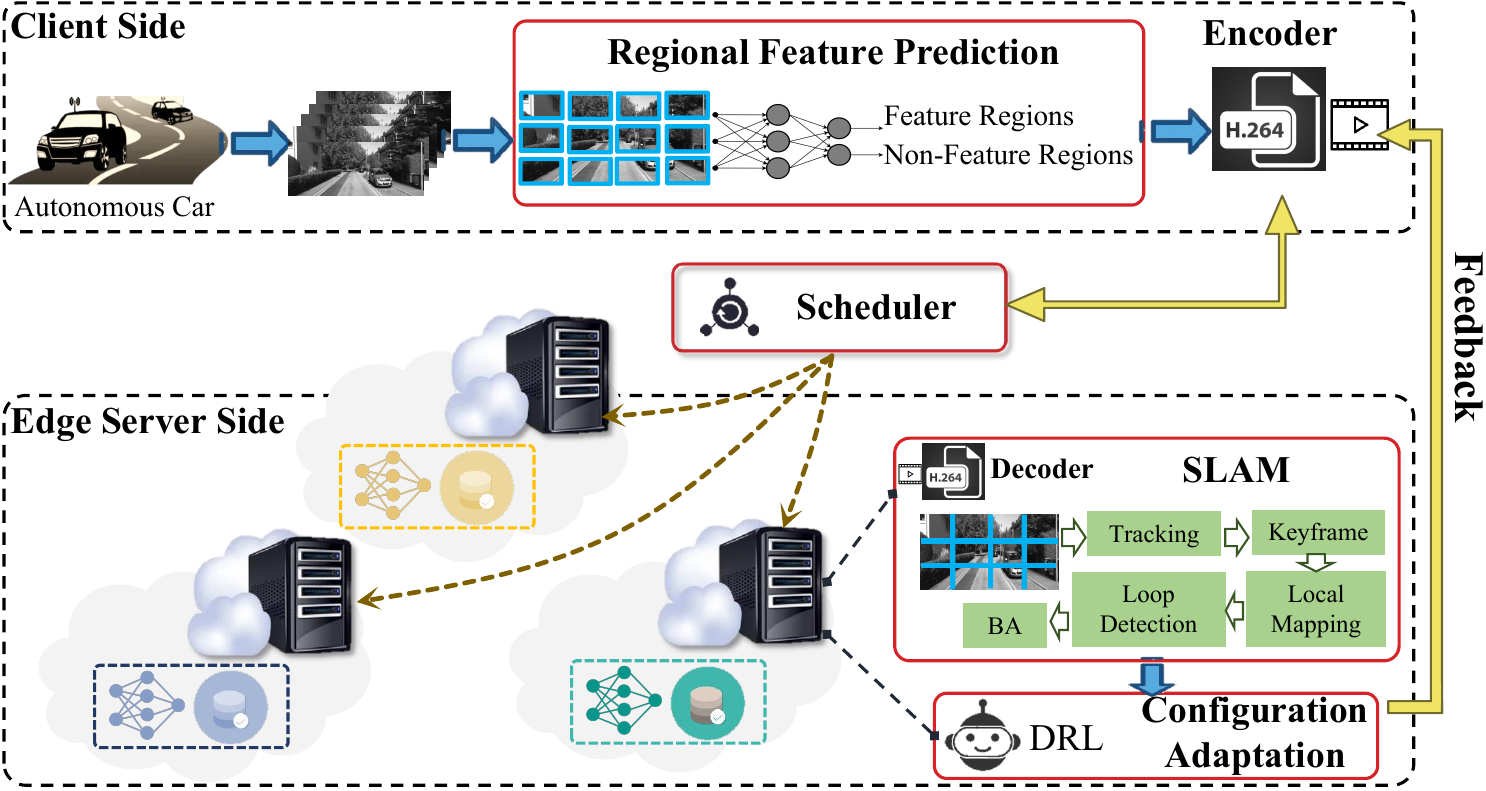}
\caption{{Proposed edge-assisted SLAM architecture.}}
\label{systemarchitecture}
\vspace{-0.5cm}
\end{figure}

{Despite its many practical applications, the general-purpose adoption of SLAM on mobile robotic systems equipped with limited computational resources is hindered by the following two key technical challenges.} \emph{Accumulated Localization Errors:} SLAM estimates the sequential movement of robotic systems, which inherently involves a certain degree of error. This error accumulates over time, leading to significant deviations from the actual values. 
\emph{High Computational Costs:} The operation of SLAM relies on frequent execution of image processing or point cloud matching. In addition, SLAM typically includes modules necessary for 
optimization calculations such as loop closure and bundle adjustment. {As such, SLAM incurs high computational costs to achieve accurate localization and mapping.} One possible solution for reducing accumulated localization errors is the construction of pose graphs in which the landmarks from previously visited places are recorded. {However, the extraction of informative characteristics (feature points) from environments during the operation of SLAM is a requirement for the reliable utilization of landmarks, which inevitably increases the computational overhead.} The countermeasure to mitigate the negative impact of high computational costs is to execute different SLAM modules in parallel, e.g., using multicore CPUs and embedded GPUs. Local parallelization can indeed help speed up the overall computations but cannot alleviate the computational burden. 


A viable approach for alleviating the computation burden induced by SLAM is to offload partial computations to edge servers that possess abundant computing resources. This has motivated recent investigations into the realm of edge-assisted SLAM, such as \cite{braud2020multipath, ahmad2020carmap, xu2020edge, ben2020edge, liu2021edgesharing, xu2022swarmmap}. 
These efforts provide a multitude of insightful contributions by addressing various challenging aspects, including the development of a multipath offloading strategy \cite{braud2020multipath}, creation of lean representations for vision data \cite{ahmad2020carmap}, extraction of semantic information \cite{xu2020edge}, task decomposition and implementation \cite{ben2020edge}, edge-assisted information sharing \cite{liu2021edgesharing}, and collaborative SLAM techniques \cite{xu2022swarmmap}.

However, the lack of attention to address two crucial issues hinders the 
practical deployment of edge SLAM in real-world scenarios.

\begin{itemize}
\item \textbf{Elasticity of Edge SLAM.} 
The offloading demand of mobile robotic systems changes over time. The existing single-server edge SLAM, i.e., edge servers independently process offloaded computations, cannot meet the demand at peak hours. In the case of autonomous driving, road congestion in a particular area leads to a sudden increase in the demand for offloading SLAM computations to the corresponding edge server. Similarly, crowd monitoring in urban areas utilizing numerous Unmanned Aerial Vehicles (UAVs) \cite{al2017crowd} necessitates consistent computation offloading and thus varies the computational load for edge servers due to the spatially dynamic nature of monitoring areas. How to improve the elasticity of edge services via inter-server cooperation for edge SLAM remains an open problem. Although inter-server cooperation has been a traditional focus in the field of edge computing \cite{zhang2023hybrid}, the intricate nature of task relationships in SLAM, such as task dependency and task parallelization \cite{ben2020edge}, presents a considerable challenge. 

\item \textbf{{Implicit and Time-Varying Requirements of Clients.}} 
Most of the existing works for edge SLAM focus on how to optimize the system for given user requirements. For example, 
\emph{Edge-SLAM} in \cite{ben2020edge} firstly focuses on the task decomposition of SLAM for edge computing based on two collected datasets and each of them comprises a collection of approximately $30$ Frames Per Second (FPS). However, in real-world scenarios, the requirements of users are usually heterogeneous in terms of task processing quality and time. 
For example, users may have different expectations of processing time when using different networks, such as WiFi, and cellular networks. The entertainment applications\footnote{https://stlpartners.com/articles/edge-computing/5g-edge-ar-vr-use-cases/}, e.g., Augmented Reality or Virtual Reality, have a higher tolerance for processing quality and latency than the HD map used for autonomous driving and navigation used for UAVs. Overall, these requirements vary across users, applications, and network conditions, consequently impacting the edge SLAM system in an unknown and time-varying way.
\end{itemize}
 
{To address the issues above, we develop a new edge-assisted SLAM architecture by orchestrating joint offloading and scheduling of SLAM tasks in multi-server scenarios.} The objective is to improve the processing efficiency of SLAM and simultaneously meet the implicit and time-varying requirements of users by exploiting both user-edge and inter-server collaboration. We take Deep Reinforcement Learning (DRL) as a fundamental tool to accomplish the complex decision-making process in the new architecture. DRL can outperform traditional heuristic algorithms by learning knowledge from effective interactions between agents and actual environments, without relying on unreasonable assumptions.
Nevertheless, edge-assisted SLAM is a representative input-driven system in which input data and network conditions comprise an exogenous and stochastic input process. In such an input-driven system, traditional state-dependent DRL algorithms suffer from high state dynamics, leading to inherent bias in policy gradient estimates and increased gradient variance during training \cite{mao2019variance, wu2018variance}. To resolve this issue, we design a new input-dependent learning framework for edge-assisted SLAM in the multi-server scenario.

{
Specifically, our journey toward the new edge-assisted SLAM architecture is structured in three steps.} First, we focus on module-wise innovation to cater both single-server and multi-server edge computing. As shown in Fig.~\ref{systemarchitecture}, the new architecture comprises five components, including \emph{regional feature prediction}, \emph{encoder/decoder}, \emph{SLAM processing pipeline}, \emph{configuration adaptation}, and \emph{scheduler}. 
Second, to improve adaptive offloading, we specifically explore optimal configurations for \emph{regional feature prediction}, \emph{encoder/decoder}, \emph{configuration adaptation}. Third, a plug-and-play module \emph{scheduler} is proposed to achieve inter-server cooperation and thus accommodate implicit and time-varying users' requirements while maintaining a balanced workload across servers.

\begin{enumerate}
    \item \emph{Architecture:} 
    Conceptually, this is the first work that incorporates inter-server cooperation as a new optimization dimension into the design space of edge-assisted SLAM for better performance-resource trade-offs on mobile robotic systems. The newly proposed architecture pushes the limit beyond existing edge-assist SLAM in which servers independently process offloaded computations.
    \item \emph{Algorithms}: {Technically, we devise module-wise innovations within the new architecture, including a regional feature prediction module for importance-aware local data processing,} a DRL-based configuration adaptation policy for adaptive offloading, and an input-dependent learning framework for the scheduling algorithm in inter-server cooperation. These innovations facilitate the adaptation and elasticity of edge-assisted SLAM. The principle behind these innovations is not SLAM-specific. We also propose a modeling method to incorporate the temporal correlation of users' requirements into the scheduler.
    \item \emph{Experiments}: 
    {Using extensive experiments, we show that the new edge-assisted SLAM architecture achieves higher localization performance and lower End-to-End (E2E) latency for various network conditions.} To solidify the performance evaluation in the case of multi-server cooperation, {we conduct experiments using a newly developed simulator originally used for Spark job execution in Cloud computing.} We comprehensively evaluate the proposed input-dependent learning framework and show that the new architecture successfully satisfies the user requirements and balances the workload of servers.  
\end{enumerate}
The rest of this paper is organized as follows. In Section \ref{backgroundandsystem}, we present the proposed edge-assisted SLAM architecture. In Section \ref{algorithmdesign}, we present the details of each module in the architecture, including the design of the regional feature prediction module, the encoder/decoder, the configuration adaptation module, and the scheduler. The comprehensive performance evaluation is discussed in Section \ref{performanceevaluation}. We review the related literature in Section \ref{relatedwork} and finally conclude this paper in Section \ref{conclusion}. 

\section{System Architecture} \label{backgroundandsystem}
In this section, we introduce the details of the edge-assisted SLAM architecture by taking ORB-SLAM2 \cite{mur2017orb} as an underlying pipeline. Note that while we confine our attention to ORB-SLAM2 in this paper, the proposed edge-assisted SLAM architecture is also applicable to other vSLAM algorithms subject to minor adaptations.

\subsection{ORB-SLAM2}

ORB-SLAM2 is a popular vSLAM system that consists of three main parallel threads, including tracking, local mapping, and loop closing, as elaborated below.
 
\emph{Tracking:} The tracking thread localizes the camera with every new frame by finding features that match the constructed local map. Feature extraction and feature matching are two important modules in the tracking thread. The feature extraction module is based on the well-known ORB algorithm \cite{rublee2011orb} by detecting feature points using the FAST algorithm \cite{rosten2008faster} and assigning each of them a binary descriptor \cite{calonder2010brief}. The feature matching module then compares the feature descriptors from a new frame with those from the history frames to estimate the pose of the camera. 

\emph{Local Mapping:} Local mapping refers to the process of constructing and optimizing the feature map in a SLAM system. In ORB-SLAM2, only the eligible frames are set as keyframes to save storage and computational resources. For a new keyframe, local mapping selects the valid features, i.e., the features that occur across multiple keyframes with stable positions, as the map features to update the local map. Besides, by using a keyframe refreshing policy, local mapping ensures the continual insertion of map features in the local map while evicting redundant ones on time.

\emph{Loop Closing:} The localization performance of ORB-SLAM2 is determined by the feature matching and pose estimation results. Although ORB-SLAM2 can localize the camera in both unmapped and mapped environments, feature matching for unmapped areas may lead to accumulated drift and thus degrade the localization accuracy over time. As a result, the loop closing thread detects large loops and corrects the accumulated error in exploration. After a tracking failure, relocalization will be performed based on place recognition and pose graph optimization. 
  
\subsection{System Architecture Overview} 
Our new edge-assisted SLAM architecture differs from existing edge SLAM systems that split the SLAM processing pipeline into a client and a server module \cite{xu2020edge, ben2020edge}. 
They lead to excessive communication latency since a huge amount of keyframes are determined and transmitted. In addition, these works are SLAM-specific, i.e., their decomposition approaches are designed for a specific SLAM system. Our new architecture evaluates the tile importance within each frame on the client side and adapts the compression configurations to the importance levels and network conditions. It requires less computational and communication costs. Both the offloading and scheduling approaches are not SLAM-specific. In the following, we overview the key functionalities of each module and leave the detailed algorithm designs to Section \ref{algorithmdesign}.
  
\subsubsection{Regional Feature Prediction Module}
\begin{figure}[!t]
\centering
\subfigure[Extracted Feature Points]
{\begin{minipage}{4cm}
\centering
\includegraphics[width=1.6in]{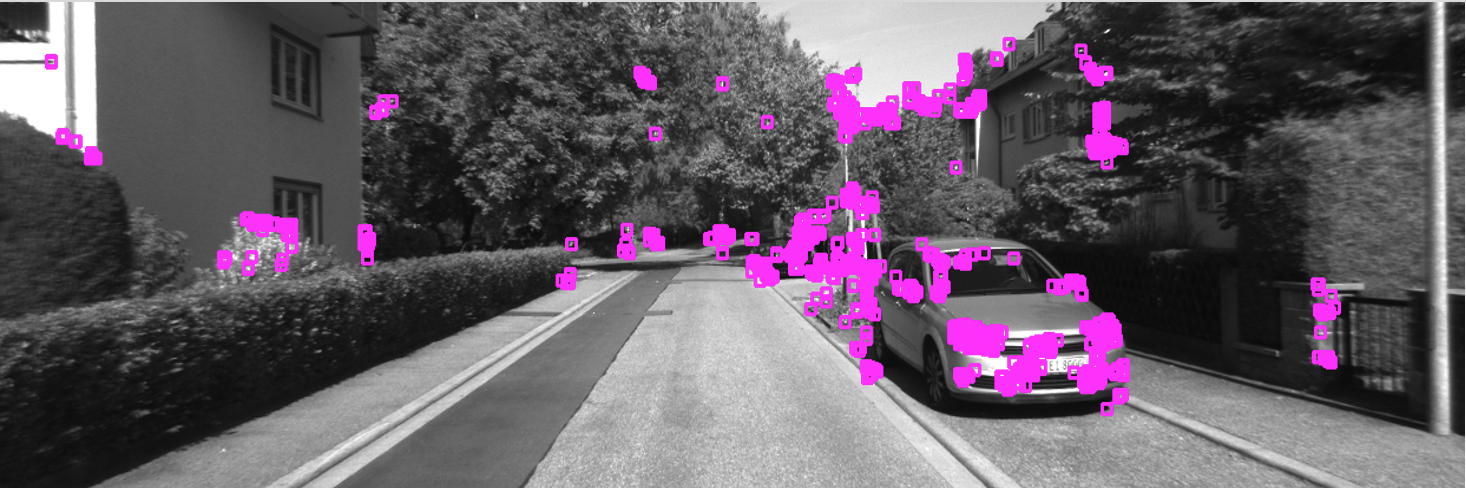}
\end{minipage}}
\subfigure[Partitioned Feature Tiles]
{\begin{minipage}{4cm}
\centering
\includegraphics[width=1.6in]{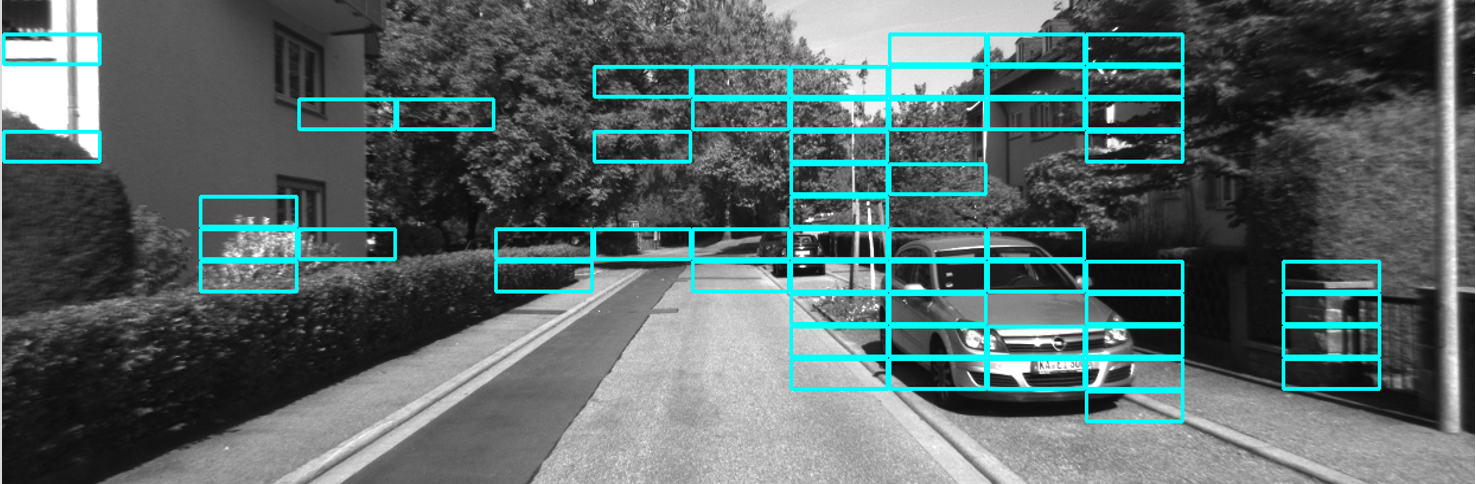}
\end{minipage}
}
\caption{A comparison of the extracted feature points and the generated tiles in a frame.}
\label{featureandregions}
\end{figure}

An important characteristic of vision data (e.g., video frames) captured by on-board cameras in mobile robotic systems is temporal correlation.
This lays the foundation of the regional feature prediction module designed for the evaluation of input video frames. Specifically, we partition each video frame into small tiles and estimate the tile importance levels according to historical knowledge. The principle of tile partition is enlightened from the ``\emph{detect + track}" framework in \cite{guo2018small, liu2019edge}, which facilitates the detection and tracking of special regions in consecutive frames. 
The scale of tile partition is based on the consideration of processing time at the client side. The partition with a small number of tiles leads to less processing time. In this case, each tile contains a large frame region, which means that more pixels that are irrelevant to SLAM processing may be involved and thus result in data redundancy in compression. On the contrary, a fine-grained partition is prone to effective compression but a large number of tiles will increase the processing time at the client side. Fig. \ref{featureandregions} shows an example of generated tiles based on the ORB feature points, where tiles containing ORB features are marked by rectangular bounding boxes. 
The feature points shown in Fig. \ref{featureandregions} are detected using the ORB feature detection algorithm proposed in \cite{rublee2011orb}. This algorithm forms the basis of the feature extraction procedure in ORB-SLAM2, which incorporates an additional step to enhance uniform feature extraction. We employ this classic algorithm to demonstrate the principle of the regional feature prediction module and also emphasize that this module is compatible with other feature-based SLAM systems. {To enable the real-time determination of the tile importance, we develop a CNN-based prediction model in the regional feature prediction module. A well-trained CNN model can immediately determine the importance levels of tiles on a newly sampled frame. If using a method based solely on the conventional ORB algorithm, the system must wait for the extraction of ORB feature points, which will extend the overall system processing time.}


\subsubsection{Encoder/Decoder}
A pair of encoder/decoder is applied to compress/decompress the tiles of a video frame. Although there are a plethora of compression configurations in the encoder/decoder module that affect the quality of frames and downstream SLAM performance, we select two key compression configurations to emphasize two types of redundancy in a video frame, namely intra-tile redundancy and inter-tile redundancy. Specifically, the intra-tile redundancy refers to the information that is irrelevant to the SLAM processing in each tile, which can be captured by the tile importance level. Based on the intra-tile redundancy, we convert the tiles into various qualities with different resolutions. The inter-tile redundancy originates from the intrinsic spatial redundancy in a video frame, which can be mitigated by using video coding techniques (e.g., H.264) that encode the tiles of a frame into a video segment. {In this way, the quantization parameter can be configured according to the tile importance levels to reduce the communication overhead. For convenience, we denote the compression configurations as a two-tuple, i.e., $<{\emph{res.}}, \emph{{QP}}>$.} {After applying compression configurations to tiles based on their importance, low-quality tiles will generate fewer feature points at the edge server. To compensate for this, more feature points will be detected in the high-quality tiles.
}
 
\subsubsection{Configuration Adaptation Module}
The configuration adaptation module adapts the compression configurations to tile importance and network conditions. 

\textbf{DRL for Configuration Adaptation:}
In an edge-assisted system, the dynamics of the input processes encompass two aspects, i.e., network conditions and vision data (i.e., tile importance). Neither of them can be predicted accurately, leading conventional adaptive compression algorithms for edge-assisted video analytics to rely heavily on heuristics and predefined assumptions \cite{galanopoulosautoml, wang2020joint, liu2025deepswarm}. DRL is committed to making adaptive decisions based on sequential interactions with the environment in a model-free way. It has been verified in many practical network environments, such as video streaming, congestion control, load balancing, and resource management \cite{mao2019park,liu2023enabling}. We leverage DRL to steer the configuration adaptation module. To deal with the challenges of training caused by huge state space in such an input-driven system, we adopt the principle of decentralized training and centralized learning architecture to enable learning efficiency. {In addition, a quality of experience (QoE) metric that balances the reduction of communication latency and improvement of the SLAM performance is adopted as the reward function for training the DRL agent to gradually learn to make better compression decisions through reinforcement. Using the QoE, the DRL agent makes decisions that effectively alleviate the local computing burden at the expense of communication resources, and also guarantee the SLAM positioning performance at the edge.}

\subsubsection{Scheduler} 
The scheduler module improves the elasticity of edge SLAM by satisfying the implicit and time-varying requirements of users.

\textbf{Constrained RL Problem:}
Our primary goal is to enable the scheduler to respect the requirements of users, which, however, is intractable since those requirements are usually implicit and time-varying. For example, the SLAM pipeline includes multiple tasks whose processing is dependent or parallel \cite{braud2020multipath}. Those tasks differ from each other in duration, complexity, and requirements, which significantly challenge the design of scheduling algorithms. Although DRL empowered by Deep Neural Network (DNN) models has the potential to learn unknown reward functions by incorporating expensive-to-evaluate objectives, such as heterogeneous task attributes and users' requirements, it is not robust enough as the changing of users and environments with different input data needs a re-training of the DRL model. 

By adopting the idea that encoding expensive-to-evaluate elements into constraints is robust to changing environments \cite{galanopoulosautoml}\cite{lindner2022interactively}, we propose to characterize the scheduling problem as a constrained RL problem by constructing a reward function with easy-to-evaluate metrics and a constraint with unknown and expensive-to-evaluate users' requirements.


\textbf{Motivation of Input-Dependent Gaussian Process (GP):} 
As an input-driven system, the stochastic input sequences of edge-assisted SLAM cause a considerable impact on the state dynamics and rewards, which then incur high variance during the training of policy gradient methods \cite{mao2019variance, wu2018variance}. Traditional DRL algorithms, such as baseline subtraction by the value function in Advantage Actor Critic (A2C), cannot alleviate the impact. This is because subtracting a baseline from the expected return based on a specific state may offer limited information for constraint violation. For example, a favorable action may yield a poor return when faced with an unfavorable state, while conversely, an unfavorable action can result in a high return given an advantageous input sequence. Both cases aforementioned cannot avoid constraint violation during the training of DRL agents.

Cathy et al. \cite{wu2018variance} derives an action-dependent baseline while Hongzi et al. \cite{mao2019variance} proposes an input-dependent baseline, both of them are used to replace the traditional value function-based baselines. 
Enlightened by them, we propose an input-dependent GP as a new kind of baseline for variance reduction which fully captures the temporal correlation of constraints from Markov Decision Process (MDP) trajectories. 
The key idea is to avoid states falling into regions in which the agent will violate constraints. To achieve that, we design an input-dependent GP model to estimate whether a state belongs to the safe regions or not by fitting a GP model with the information related to both state and input sequences. As such, the state transition of MDP could be guided by the input-dependent GP model, i.e., the state transition with a high risk of constraint violation will be penalized. 

In summary, the advantages of the newly proposed input-dependent learning framework are twofold:

\emph{Efficient and Cost-Effective:} The input-dependent GP is cost-effective as there are no extra training processes and complex neural networks for the fitting of a GP model. It is an efficient model for the constrained RL problem with expensive-to-evaluate constraints as the fitted GP can capture temporal features from high-dimensional information. 

\emph{Scalable:} The DRL policy trained with the guidance of the input-dependent GP is scalable as it can be transferred to different users and environments in an online way. For a new user, the GP model could be fitted using the new MDP trajectories with collected posterior information about constraint violation and then used to fine-tune the DRL policy. 


\section{Algorithm Design} \label{algorithmdesign}
In this section, we introduce the detailed algorithm designs for edge-assisted SLAM.  
 
\subsection{Regional Feature Prediction}
To enable accurate and real-time tile-oriented importance evaluation, we develop a regional feature prediction model by adopting the principle of lifelong learning \cite{feng2020livedeep, feng2021liveobj}. In the edge-assisted SLAM system, since video frames and the corresponding tiles are generated on the fly, lifelong learning enables the prediction model to be updated in an online way, thereby maintaining its prediction performance. 
   
\subsubsection{Workflow of Regional Feature Prediction}
To enable tile-oriented evaluation on resource-limited devices, our key idea is to take a lightweight NN model as the backbone network of the prediction model and modify the output units of the last layer according to the pre-defined tile importance levels. We aim to train the model for capturing the natural inter-frame and intra-frame correlation in video frames and thus the lightweight prediction could be achieved.

Particularly, we exploit a \emph{sampling and prediction} strategy to guarantee fast training and real-time prediction in the prediction model. To achieve this goal, we refine the processing pipeline in the regional feature prediction module by managing a local buffer that saves $k$ sampled frames from a time slot. These sampled frames are taken as the basic processing units in the prediction model to perform the inference procedure and generate training samples.
 
The workflow of the regional feature prediction is shown in Fig. \ref{predictionmodel}, where the prediction model starts the inference on a new sampled frame. Specifically, for the inference pipeline, the sampled frame in the buffer is first partitioned into $x*y$ small tiles, which then be passed into the prediction model to predict their importance levels. The 
prediction results will be applied to the remaining frames based on the temporal correlation until the arrival of the next sampled frame. To generate training samples, the sampled frame is also used to extract features by using the ORB algorithm \cite{rublee2011orb}. Based on the feature extraction results, the real importance levels for those tiles are obtained. {In this way, the training samples containing in total $k{\times}x{\times}y$ training tiles can be formed from all the sampled frames in a buffer. Both the tile number and downsample interval are taken as hyperparameters in the system implementation.} 

 
\subsubsection{Training Procedure}
\begin{figure}[!t]
\centering
\includegraphics[width=3.4in]{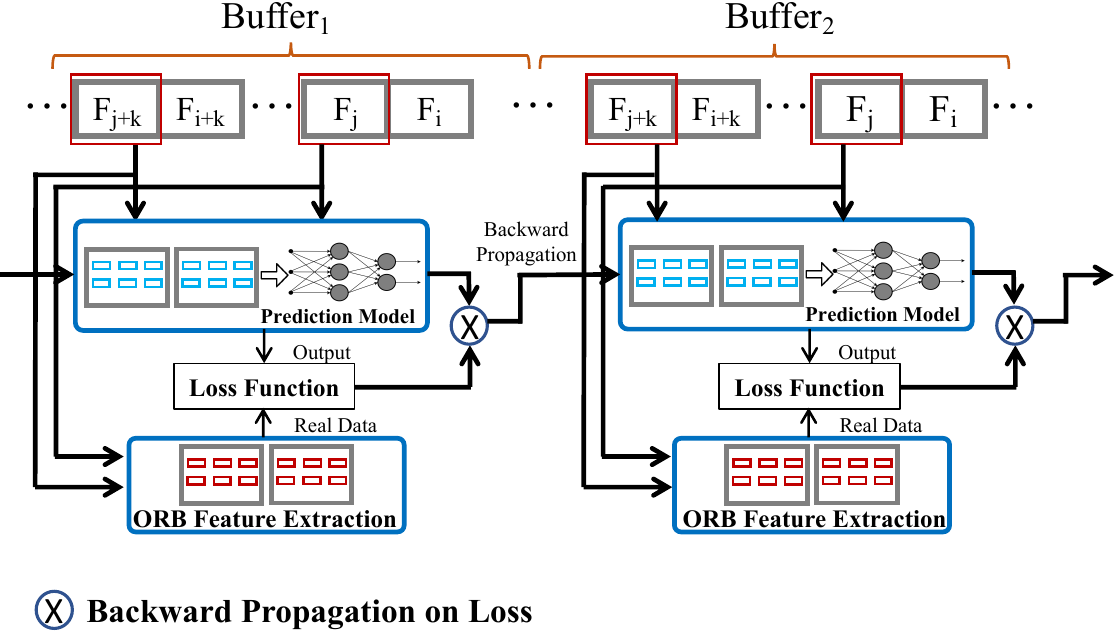}
\caption{Workflow of the regional feature prediction module.}
\label{predictionmodel}
\vspace{-0.4cm}
\end{figure}

With the frame updating in the local buffer, the prediction model is trained at runtime based on the collected training samples. The training procedure is also presented in Fig. \ref{predictionmodel}, where \emph{Buffer{$_1$}} and \emph{Buffer{$_2$}} represent the local buffer storing sampled frames from two consecutive time slots, namely $t_1$ and $t_2$, respectively. 
   
For a new sampled frame, upon acquiring the actual importance-related information of its tiles through ORB feature extraction, the loss function will be calculated and then the prediction model will be updated by the way of backward propagation. The updated prediction model based on the sampled frames in \emph{Buffer{$_1$}} will be used to infer the sampled frames in 
the \emph{Buffer{$_2$}}.  
 
Recall that the inference procedure in the prediction model is triggered by the next sampled frame. To enable real-time inference on the sampled frames in the \emph{Buffer{$_2$}} of Fig. \ref{predictionmodel}, we complete the updating of the prediction model before the end of \emph{Buffer{$_1$}}. To achieve this goal, we constrain the training procedure at each buffer by customizing the minimum loss value, number of training epochs, and batch size. It means that in the system implementation, both the inference and training procedures are performed in parallel in the regional feature prediction module.

Importantly, the backbone network in the prediction model can be substituted with different pre-trained neural network models, depending on the desired accuracy and the available onboard computational resources. In our experiments, we choose LeNet \cite{lecun1998gradient} as the backbone network due to its successful implementation on memory-constrained IoT devices, as demonstrated by TinyNet \cite{valdenegro2017real} in low-power embedded systems of autonomous underwater vehicles.
 
\subsection{Input-Dependent Learning for Edge-Assisted SLAM} \label{decisiona3c}

\subsubsection{MDP Formulation}
In the new edge-assisted SLAM architecture, configuration adaptation and scheduling constitute two distinct steps. 

We first take the configuration adaptation as a sequential decision-making problem and define a QoE to characterize the communication overhead and SLAM performance. The QoE can be used to constitute the reward function as it reflects the trade-off of system costs and benefits in the edge-assisted SLAM system. 

Specifically, let $a_{t}$ represent the compression configuration applied to the video frame at time $t$, $t=1,..N$. The obtained data size after applying $a_{t}$ is $Q(a_{t})$. The instantaneous bandwidth at time $t$ is denoted as $B_t$. Let $P_{t}$ denote the performance metric of SLAM processing. To reflect the prediction performance, we define $P_{t}$ as the difference between the predicted tile importance and that obtained via ORB feature extraction at the edge side. This is because ORB feature extraction is a key step in SLAM and the extracted map points determine the downstream mapping and localization performance. The QoE is thus defined as: 
\begin{equation}
\centering
\label{qoedefinition}
QoE = \alpha * \sum_{i=1}^N P_{i} - \beta * \sum_{i=1}^N \frac{Q(a_{i})}{B_i},
\end{equation}
where $\alpha$ and $\beta$ are weight factors used to strike a balance between bandwidth consumption and SLAM performance.  

In the formulation of MDP, we denote the state, action, and reward at time $t$ as $s_t$, $a_t$, and $r_t$, respectively. 
Specifically, when the DRL agent observes state $s_t$, it decides action $a_t$ based on a control policy. After acting $a_t$, a reward signal $r_{t+1}$ will be obtained to optimize the control policy and the state switches to $s_{t+1}$. 
Based on the Markovian property of the sequential configuration adaptation process above, the MDP is defined as:
\begin{equation}
\centering
\label{mdp}
M_{config} = (S_{t}, A_{t}, R_{t}, \pi_{t}),
\end{equation}
where $\pi_{t}$ refers to the control policy at time $t$. We next elaborate on the definition of these terms and illustrate the training of DRL.


\subsubsection{Configuration Adaptation Based on A3C}
\ 
\newline
\indent
\textbf{State Space:} An exhaustive state space enables the DRL agent to comprehensively learn the optimal control policy. To provide sufficient information that reflects the stochastic input sequence, we incorporate four environmental elements into the state space, i.e., \emph{tile importance}, \emph{throughput}, \emph{latency}, and \emph{last configuration}, denoted as $s_t = ({e_t}, \vec{b}, \vec{l}, {a_t})$ for the time $t$. Particularly, to provide more bandwidth-related information, we further integrate $k$ past measurements in both throughput and latency, i.e., $\vec{b}=\{b_{t-k+1}, b_{t-k+2},...b_{t}\}$ and $\vec{l}=\{l_{t-k+1}, l_{t-k+2},...l_{t}\}$. 
\textbf{Action Space:} The action space corresponds to the available choices of $a_t$, representing the compression configurations in the edge-assisted SLAM system. As we incorporate the resolution (\emph{res.}) and the quantization parameter (\emph{QP}) into the compression configurations, $a_t$ is denoted by a two-tuple, i.e., $a_t = \{res._t, QP_t\}$.
  
\textbf{Control Policy:} While the action space is bounded in the edge-assisted SLAM system, the state space tends to be infinite because of the unpredictable and stochastic bandwidth variations, resulting in a rich variety of $(s_t, a_t)$. To effectively evaluate the state-action pairs, we employ a neural network as the control policy. We adopt the policy gradient method from \cite{sutton2000policy} to maximize the expectation of the cumulative reward, i.e., $\bold{E}(\sum_{t=1}^{t=N}{\gamma}^tr_t)$, where $\gamma$ is a discount factor for the future QoE. 

The input-driven system significantly challenges the training efficiency of DRL algorithms \cite{mao2019park}\cite{mao2019variance}\cite{han2021tailored}\cite{mao2019learning}\cite{zhu2021network}. To address that, we adopt the architecture of decentralized training and centralized learning \cite{sutton2018reinforcement}, forming an Asynchronous Advantage Actor-Critic (A3C) algorithm as the training mechanism of the DRL agent. The A3C algorithm employs two neural networks: One for the control policy $\pi_t$ and the other for evaluating $\pi_t$. Specifically, the neural network responsible for the control policy $\pi$ is referred to as the policy network, which is defined by the network parameter $\theta_p$. The critic network evaluates $\pi$ by updating its network parameter $\theta_v$ based on a value function. In addition, the critic network evaluates $\pi$ by updating its network parameter $\theta_v$ through a value function.
 
The updating rule of $\theta_p$ is:
\begin{equation}
\centering
\label{policyupdating}
\theta_p \leftarrow \theta_p + l  \sum_{t=1}^{T}\Delta_{\theta_p}log{\pi_{\theta_p}(s_t, a_t)A_{\theta_p}(s_t, a_t)},
\end{equation}
where $l$ is the learning rate. $A_{\theta_p}(s_t, a_t)$ is the advantage function to evaluate the expected reward obtained by following the control policy $\pi_{\theta_p}$. That is, the relationship between the policy network and the critic network is characterized by $A_{\theta_p}(s_t, a_t)$.
 
For the critic network, it uses the value function $v^{\pi_{\theta_p}}$ to evaluate the control policy. With the estimated value function $V^{\pi_{\theta_p}}$, the advantage function is calculated as:
\begin{equation}
\centering
\label{advantagefunc}   
A_{\theta_p}(s_t, a_t)=r_t+\gamma{V^{\pi_{\theta_p}}(s_{t+1;{\theta_v}})-V^{\pi_{\theta_p}}(s_{t;{\theta_v}})}.
\end{equation}
The updating of $\theta_v$ is shown in (\ref{criticupdating}) \cite{konda2000actor}:
\begin{equation}
\centering
\label{criticupdating}  
\theta_v \leftarrow \theta_v + \alpha{'}  \sum_{t=1}^{T}\Delta_{\theta_v}(r_t+\gamma{V^{\pi_{\theta_p}}(s_{t+1; {\theta_v}})-V^{\pi_{\theta_p}}(s_{t; {\theta_v}})})^2.
\end{equation}

\textbf{Parallel Training:}
During the training of the DRL agent, we improve the efficiency of experimental exploration by employing multiple threads, with each thread managing an agent with an individual MDP. In this parallel learning way, the decentralized agents produce diverse training tuples $\{\emph{state, action, reward, next state}\}$ synchronously by experiencing different environments. The centralized agent can consequently achieve faster actor-critic training by updating the control policy using those tuples. Subsequently, the model parameters of the newest control policy will be forwarded to the individual threads. 

\subsubsection{Input-Dependent Learning for Scheduling}
In this part, we formally formulate the constrained DRL problem and then present the newly developed scheduling algorithm under the input-dependent learning framework.


\textbf{Formulation of Constrained RL Problem:} Recall that the scheduler in Fig. \ref{systemarchitecture} is responsible for dispatching tasks (compressed frames) from the requesting user to different servers. The primary objective is to reduce the processing time of sub-tasks and also balance workload among servers.
The action is set as $\dot{a}_t=\{1,2,...K\}$, determining which server the task at time $t$ will be dispatched to. The state observed at time $t$, denoted as $\dot{s}_t$, encompasses task patterns, link conditions, and queue states of servers. Specifically, task patterns are specified by the size of each incoming frame, which will determine the sizes of sub-tasks in the SLAM pipeline. Link conditions include both instantaneous latency and throughput of the corresponding link between the user and the server. The queue state represents the number of sub-tasks accumulated in the queue of each server. 
 
The definition of reward function considers both the processing performance of sub-tasks and workload balancing of servers. The former item is denoted by the fraction of sub-tasks that surpass their deadlines and active sub-tasks in each queue while the latter item is determined by calculating the standard deviation of the queue length across all queues. The formal definition is:
\begin{equation}
\centering
\label{rewardfunctionofscheduling}
R_{sched.} = \exp({ST_{ddl} / ST_{a}})+1/(1+\exp(-\delta_{queue})),
\end{equation}
where ${ST_{ddl}}$ and $ST_{a}$ denote the number of sub-tasks that surpass their deadlines and the sum of active sub-tasks in each queue. $\delta_{queue}$ represents the standard deviation of queue length. The expectation operation aims to normalize the items to a unified range.

The constrained RL problem is then formulated as:
\begin{equation}
\begin{aligned} \centering
\label{constrainedRL}
&\dot{a}^*_t=\mathop{\arg\min}_{\dot{a}_t\in{\chi}} {\rm \lim_{T \to \infty}}\frac{1}{T}{\sum_{t=1}^T}\gamma^tR_{sched.}
(t),\\
&s.t. {\,}{w}_j \leq DDL_j, \mathbf{for}{\,}j \in{J_t},\\
\end{aligned}
\end{equation}
where $\chi$ is the action space that involves feasible actions, $T$ is the time horizon, ${J_t}$ denotes the sub-tasks in the SLAM processing pipeline at time $t$, and ${w}_j$ is the processing time for sub-task $j$. $DDL_j$ is the desired processing deadline of sub-task $j$ and is unknown to the system. 
 
\textbf{Definition of Input-Dependent GP:}
The input-dependent GP is constructed by employing a GP to model the unknown constraints based on past observations of input processes and system states. Specifically, the observations are divided into two parts, i.e., the system inputs $\mathbf{X}\cong\{\dot{s}_{\tau}, \dot{a}_{\tau}\}_{\tau=1}^T$ and the class labels $\mathbf{y}\cong\{\mathbbm{1}_{\dot{s}_{\tau+1}}\}_{\tau=1}^{T}$ for a time horizon of $T$, where $\mathbbm{1}_{{\dot{s}}_{t+1}}$ is a binary indicator for system states. As such, we use the input-dependent GP to achieve GP prediction about the system states. 

Notably, the key idea of GP prediction is to squash the latent 
function $f(\mathbf{X})$ and then obtain a prior on $p(\mathbf{y}|
\mathbf{X})=\sigma(f\mathbf{(X)})$ \cite{williams2006gaussian}. The prediction distribution to a vector $X_*$ can be obtained as:
\begin{equation}
\label{predictionprob_0}
\centering
p(f_*|\mathbf{X},\mathbf{y},X_*)=\int p(f_*|\mathbf{X},X_*,\mathbf{f})p(\mathbf{f}|X_*,\mathbf{y})d\mathbf{f}.
\end{equation}
The input-dependent GP model is a model-free method that requires 
only sufficient observations to fit the kernel function
$\rho(\mathbf{X}; \mathbf{X'})$. The fitted kernel that captures 
historical correlation is then used to construct the function at any vector $X_*$, i.e., $f_*$. 

We first present the fitting process of GP by updating the mean and variance of $f(\mathbf{X})$ as:
\begin{equation}
\centering
\label{meanvalue_gp}
\mu_{f,t}(\mathbf{X})=\mathbf{k}_t(\mathbf{X})^T(\mathbf{K}_t+\sigma^2I)^{-1}\mathcal{Y}_t,
\end{equation}
and 
\begin{equation}
\centering
\label{variance_gp}
k_{f,t}(\mathbf{X},\mathbf{X}')=\rho(\mathbf{X},\mathbf{X}')-
\mathbf{K}_t(\mathbf{X})^T(\mathbf{K}_t+\sigma^2\mathbf{I})
\mathbf{k}_t(\mathbf{X}').
\end{equation}
 
The posterior $p(\mathbf{f}|X_*,\mathbf{y})$ in (\ref{predictionprob_0}) can be approximated by a Gaussian distribution $q(\mathbf{f}|X_*,\mathbf{y})$, thus forming the Laplace approximation with a theoretical guarantee according to \cite{williams2006gaussian}. With the fitted GP, the posterior function $f*$ can be constructed, where its mean and variance is $E_q[f_*|\mathbf{X},\mathbf{y},X_*]$ and $V_q[f_*|\mathbf{X},\mathbf{y},X_*]$, respectively, under the Laplace approximation. 
The prediction based on fitted GP is then:
\begin{equation}
\label{predictionoverfittedGP}
\centering
\hat{\dot{s}}_*:=\int \sigma(f_*)q(f_*|\mathbf{X},\mathbf{y},X_*).
\end{equation}
For the details of the Laplace approximation, readers are referred to Section 3.4 in Williams' book \cite{williams2006gaussian}. We adopt the Radial Basis Function (RBF) as the basic kernel function in experiments.

\textbf{Modified MDP with Theoretical Analysis:}
By extending (\ref{advantagefunc}), a new advantage function 
that incorporates the input-dependent GP is defined as: 
\begin{equation}
\begin{aligned}
\centering
\label{advantagefunctionwithGP}
A_{\dot{\theta_p}}(\dot{s}_t, \dot{a}_t, \hat{\dot{s}}_{t+1})=&r_t(\dot{a}_{t},\dot{s}_{t},\hat{\dot{s}}_{t+1})+{\gamma}V^{\pi_{\dot{\theta}_p}}(\dot{s}_{t+1;{\theta_v}},\hat{\dot{s}}_{t+1})\\
&-V^{\pi_{\dot{\theta_p}}}(\dot{s}_{t;{\dot{\theta}_v}}).
\end{aligned}
\end{equation}
Notably, the estimated $\hat{\dot{s}}_{t+1}$ is used to indicate whether the constraint will be violated (unsafe) or upheld (safe).  
   
It can be seen that the new reward function in (\ref{advantagefunctionwithGP}) also embeds $\hat{\dot{s}}_{t+1}$, defined:
\begin{equation}
\begin{aligned}
\centering
\label{newreward}
r_t(\dot{a}_{t},\dot{s}_{t},\hat{\dot{s}}_{t+1})=r_t(\dot{a}_{t},\dot{s}_{t})+\mathcal{P}\mathbbm{1}_{\hat{\dot{s}}_{t+1}},
\end{aligned}
\end{equation}
where $r_t(\dot{a}_{t},\dot{s}_{t})$ represents the scheduling reward (\ref{rewardfunctionofscheduling}).
$\mathcal{P}$ is the penalty term used to penalize the agent for its unsafe state transition and is defined as:
\begin{equation}
\begin{aligned}
\centering
\label{penalty}
\mathcal{P}={\exp}\left(\sum_{j=1}^J(\max(0.001,\frac{w_j-DDL_j}{DDL_j}))\right),
\end{aligned}
\end{equation}
where $0.001$ is a manually set value. 
 
According to the new definition of both the advantage function and reward function defined above, the proposed input-dependent GP confines the MDP to a reduced transition space, leading to the new definition of MDP as well as the theorem of optimal scheduling.

\begin{myDef}
\label{definition1}
Given a MDP $M_{sche.}=(S, A, P, R, \pi_{sche.})$ and the unsafe states $S^-\subset{S}$, the new MDP is defined as 
$M^-_{sche.}=(S^-, A, P^-, R^-, \pi^-_{sche.})$, where $P^-=P(\mathbbm{1}_{\hat{\dot{s}}_{t+1}}), \hat{\dot{s}}_{t+1}
\in{S^-}$ and $R^-$ is a function of $\dot{a}_t$, $\dot{s}_t$ and $\hat{\dot{s}}_{t+1}$. 
\end{myDef}
  
\begin{theorem}
\label{theorem1}
Considering the optimal policy of $M_{sche.}$ and $M^-_{sche.}$ is $\pi_{sche.}$ and $\pi^-_{sche.}$, respectively, if states under $\pi^-_{sche.}$ never fall into the unsafe state region ${S^-}$, $\pi^-_{sche.}$ is the optimal policy of $M_{sche.}$, i.e., $\pi^*_{sche.}=\pi^-_{sche.}$. 
\end{theorem}
 
Theorem \ref{theorem1} illustrates the condition that achieves the optimal policy $\pi^*_{sche.}$, which means that $\pi^*_{sche.}$ could be obtained by DRL training after $P^-=0$. That is, the optimal scheduling policy highly relies on the fitting performance of the input-dependent GP, which is usually depicted by the convergence according to \cite{williams2006gaussian}. The training process of the scheduling algorithm is shown in Algorithm \ref{trainingscheduling}, which includes the fitting process of the input-dependent GP model. We then use ablation experiments to evaluate the impact of the input-dependent GP on the training and scheduling performance of Algorithm \ref{trainingscheduling}. 

\begin{algorithm}[t]
\caption{Training and Scheduling}
\label{trainingscheduling}
\SetAlgoLined
\KwIn{$GP_f=0$}
    \begin{algorithmic}[1]
    \FOR{$t$=1,2,..,$T$}{
        \STATE  For frame and system at $t$, observe $\dot{s}_t$;
        \IF {$1<t\leq{T_0}$}
        \STATE \textbf{Collects} $\mathbf{X}_{t-1}$, $\mathbf{y}_{t}$;
        \STATE \textbf{Fit} GP using (\ref{meanvalue_gp}) and (\ref{variance_gp});
        \ELSIF{$t>T_0$}
        \STATE{$GP_f=1$;}
        \IF {$t{\,}\%{\,}{T_1}==0$}
        {\STATE \textbf{Updates} GP using (\ref{meanvalue_gp}) and (\ref{variance_gp});   }
        \ENDIF
        \ENDIF
        \STATE \textbf{Obtains} $\dot{a}^*_t$ through policy $\pi_t$;
        \IF {$GP_f==1$}
        \STATE \textbf{Estimates} $\hat{\dot{s}}_{t+1}$ though (\ref{predictionoverfittedGP}) by setting $X_*$ with $\dot{s}_t$ and $\dot{a}^*_t$;
        \IF {$\hat{\dot{s}}_{t+1}$ is safe}
        \STATE \textbf{Takes} $r_t(\dot{a}_{t},\dot{s}_{t})$ as reward;
        \ELSE
        \STATE \textbf{Takes} $r_t(\dot{a}_{t},\dot{s}_{t},\hat{\dot{s}}_{t+1})$ as reward;
        \ENDIF
        \STATE \textbf{Updates} $\pi_t$ using (\ref{policyupdating}), (\ref{advantagefunctionwithGP}), (\ref{criticupdating}) under the parallel training architecture;
        \ELSE
        \STATE \textbf{Takes} $r_t(\dot{a}_{t},\dot{s}_{t})$ as reward;
        \STATE \textbf{Updates} $\pi_t$ using (\ref{policyupdating}), (\ref{advantagefunc}), (\ref{criticupdating}) under the parallel training architecture;
        \ENDIF
    }
    \ENDFOR
    \end{algorithmic}
\end{algorithm}
  

\section{Performance Evaluation} \label{performanceevaluation}
In this section, we evaluate the performance of the edge-assisted SLAM architecture through two steps. First, we perform experiments in a single-server scenario for fair comparison with a classical edge-SLAM system and showcase the performance improvement caused by our architecture. 
Second, in the extended experiments, we consider a general scenario with multiple servers to verify the performance induced by all the modules.
 
\subsection{Performance of Configuration Adaptation}  
The experimental results verify that our architecture successfully reduces the end-to-end latency and improves the localization performance of SLAM. We also evaluate the training performance of A3C within the configuration adaptation module.
{
We use a hardware system to perform training, testing, and execution of the proposed architecture. The main hardware conditions that run over Ubuntu $16.04$ OS include: Intel i-$7$-$11700$ CPU, NVIDIA GeForce RTX $3060$ GPU, 1T disk and $16$GB RAM.
}

\subsubsection{Experimental Setup}
The experimental settings for each module of the new edge-assisted SLAM architecture are introduced. 

\textbf{Regional Feature Prediction:}
\label{featureprediction}
In this module, each video frame is uniformly partitioned into $9$$\times$$15$ tiles, and the sample interval is set as $k=4$. It means that at each time slot, there are in total of $4$ frames in the local buffer. As such, the batch size of the training data within each time slot is $4\times9\times15$. To simplify the implementation, we set two types of importance levels at each tile, which characterize whether ORB features will be extracted within the tile or not. We adopt the LeNet \cite{lecun1998gradient} as the backbone network of the prediction model. The original LeNet has five layers, including $2$ 2D convolutional layers and $3$ Fully-Connected (FC) layers. We modify its output layer as $2$ linear units to correspond to the importance levels. The SGD optimizer is adopted in the training and the learning rate is set as $0.001$. We use the cross entropy to calculate the training loss. 
 
\textbf{Encoder/Decoder:} The first knob in this module is resolution, which is constrained as one of five options, i.e., \emph{res.} $\in$ \{1, 0.9, 0.8, 0.7, 0.6\}. Once the resolution for frame tiles is configured, the encoder module employs the H.264 compression standard to encode them into a video segment. Video encoding is accomplished by the \emph{ffmpeg}\footnote{http://ffmpeg.org/} toolbox and the x264 codec, where the tiles will be encoded by different \emph{QP} options, set as \emph{QP} $\in$ \{20, 24, 28, 32, 36\}. Therefore, all possible combinations $<\emph{res.},  \emph{QP}>$ comprise the action space. 

During the experiments, we configure the output of the A3C policy network to correspond to different importance levels. Through the operation of the encoder, each video frame will be compressed and encoded into a video segment including different sub-segments with different qualities. 

\textbf{Configuration Adaptation:} 
In the A3C network architecture,
the input layer consists of a series of FC layers with $128$ neurons for both the policy network and the critic network. In the hidden layer of the policy network, we adopt two FC layers, with each of them having $5$ linear units that correspond to the action space. Softmax operator is utilized in the policy network to output actions in a probabilistic way.
In the critic network, the hidden layer has a $128$-FC layer along with a linear unit employed for estimating the value function. ReLu function is adopted as the activation function while the Adam optimizer is adopted as the optimizer in the A3C training. The learning rate is set to $1e-3$ while $\gamma$ is $0.99$.
We employ the parallel training architecture to train the A3C algorithm for configuration adaptation. Specifically, training is conducted using 16 parallel agents, and the training trajectories are synchronized and aggregated in a centralized agent for policy updating. Each parallel agent executes its training process by taking different experimental seeds and input sequences. 

\textbf{Datasets:} Our experiments involve two types of datasets, one for network traces and another for sensory data. To evaluate the robustness of the configuration adaptation module under various network conditions, we adopt both real-measured and simulated network traces. Specifically, we use the network traces to simulate two network scenarios, one representing network congestion and the other representing hybrid 4G/5G coexistence. The traces of the former scenario are sampled from the 3G/HSDPA \cite{riiser2013commute} that contains the practical network measurements. 

We randomly select $9000$ network samples from the 3G/HSDPA dataset and partition them into $16$ traces. Among these traces, $10$ of them are designated as the training dataset and the remaining $6$ as the test dataset. 
We further generate the system-level simulations of network traces via a network simulator from \cite{oughton2019open}. 
We develop a network environment with the coexistence of $4$G and $5$G networks and extract throughput measurements to construct the network traces. Specifically, in total $3000$ samples are generated from the simulator. We partition them into $12$ traces and $10$ of them are incorporated into the training dataset while the remaining $2$ are used as the test dataset. {For the sensory data, we use the visual odometry dataset from the popular autonomous driving dataset KITTI \cite{geiger2013vision}. There are a total of $22$ sequences in the dataset and each of them has hundreds of video frames. We select a part of video frames as the sensory data sample.}



\begin{figure*}[htbp]
{
\begin{minipage}[t]{0.3\linewidth}
\centering
\includegraphics[scale=0.38]{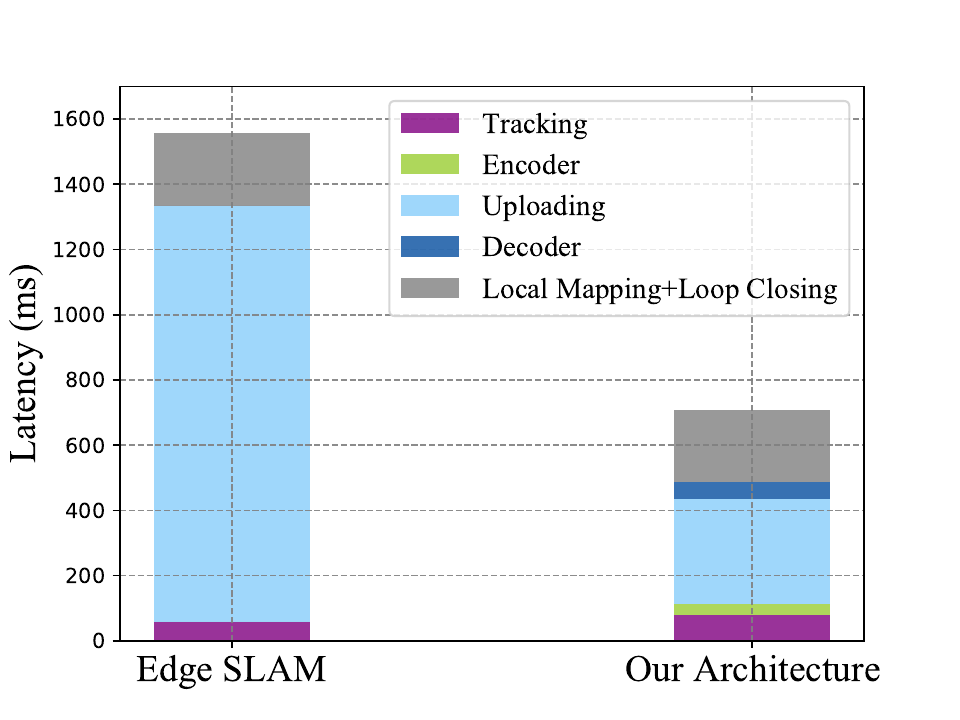}
\caption{End-to-End Latency With 3G/HSDPA Traces}
\label{systemcosthsdpa}
\end{minipage}}
\hspace{0.6cm}
{\begin{minipage}[t]{0.3\linewidth}
\centering
\includegraphics[scale=0.38]{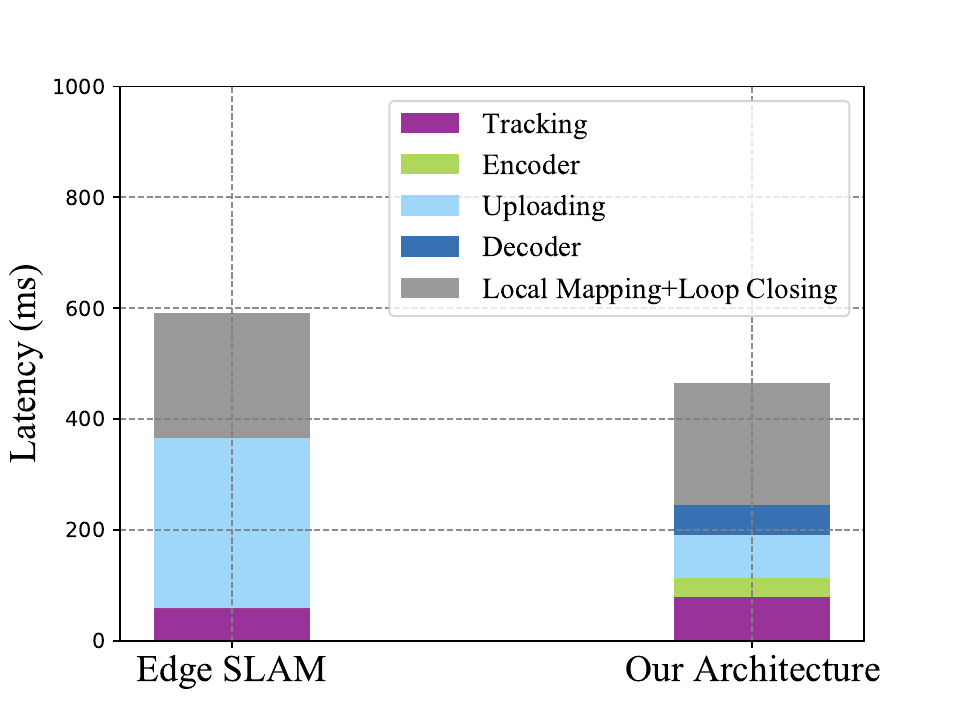}
\caption{End-to-End Latency With Simulated Network Traces}
\label{systemcostsimulation}
\end{minipage}}
\hspace{0.6cm}
{\begin{minipage}[t]{0.3\linewidth}
\centering
\includegraphics[scale=0.32]{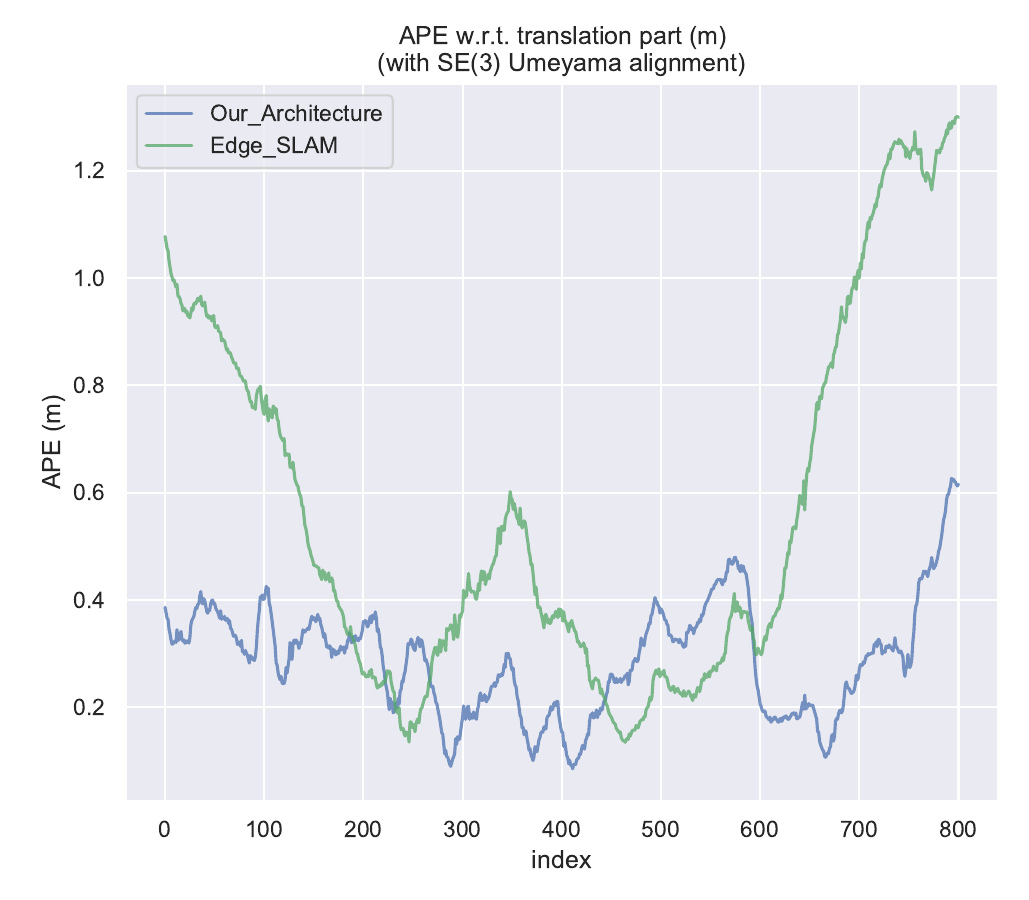}
\caption{APE Results for Frames}
\label{mappingerror_frame}  
\end{minipage}}
\vspace{-0.4cm}
\end{figure*}

\begin{figure*}
\hspace{-0.1cm}
{\begin{minipage}[t]{0.3\linewidth}
\includegraphics[scale=0.3]{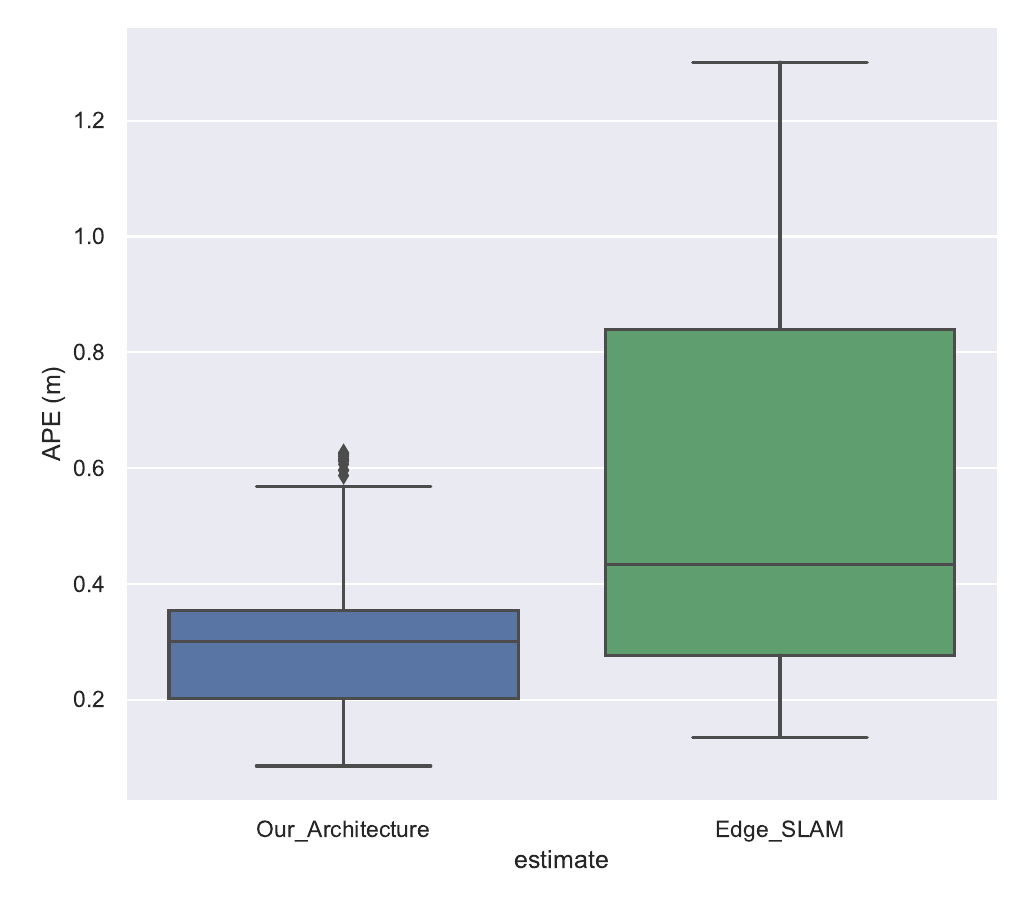}
\caption{Average APE Results}
\label{mappingerror_overall}
\end{minipage}}
  \begin{minipage}[t]{0.3\linewidth}
    \centering
    \includegraphics[scale=0.39]{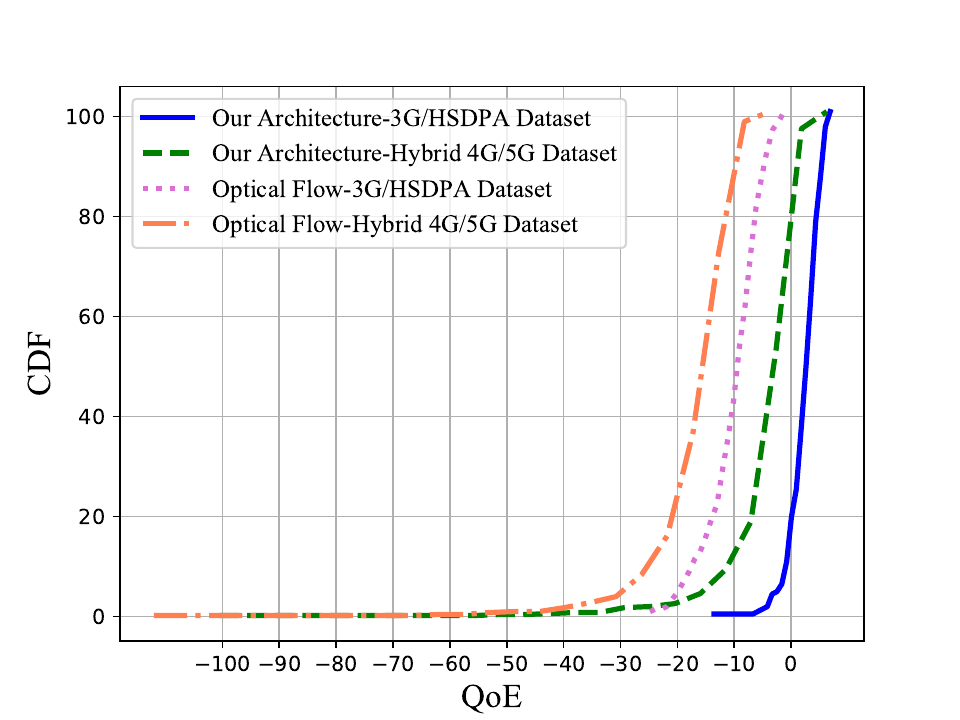}
    \caption{QoE Results}
    \label{qoekitti}
  \end{minipage}%
  \hspace{0.8cm}
  \begin{minipage}[t]{0.3\linewidth}
    \centering
    \includegraphics[scale=0.39]{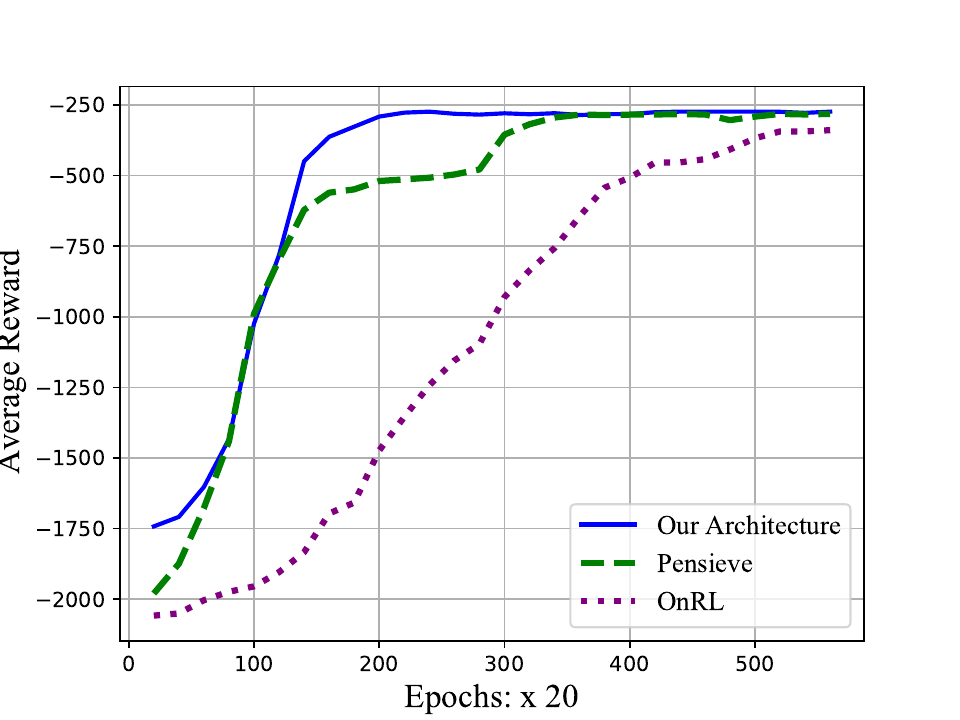}
    \caption{Comparison of Training Convergence}
    \label{trainingconvergence}
  \end{minipage}
  \vspace{-0.4cm}
\end{figure*}
 
\subsection{Results}
\emph{\textbf{E2E Latency:}} We first compare our proposed edge-assisted SLAM and a state-of-the-art edge-SLAM system (hereafter referred to as edge-SLAM) \cite{ben2020edge} in terms of E2E latency, i.e., time from the frame is generated to the last task of SLAM is processed. Both implementations are performed in a shared experimental environment on a computer. 
 
The experimental results are shown in Fig. \ref{systemcosthsdpa} and Fig. \ref{systemcostsimulation}. In Fig. \ref{systemcosthsdpa}, we depict the E2E latency for three parts: tracing, transmission, and other modules (including local mapping and loop closing). {
We omit the latency caused by the step of feedback. This is because that our system sends pose estimation results back to the client while retaining the global map on edge servers and thus 
definitely results in a lower feedback latency compared to Ali's Edge SLAM, which sends a part of map back to the client.}
For our architecture, we present the results of the main modules from the system architecture in Fig. \ref{systemarchitecture}. Moreover, the latency of the tracking module in our system is also plotted in order to make a module-level performance comparison with edge-SLAM. 
Our architecture reduces the E2E latency significantly compared with edge-SLAM, even though it introduces minimal latency due to the encoder/decoder module. In addition, the latency caused by pure SLAM pipeline in our architecture is also slightly larger than that of edge-SLAM. This is because the quality diversity within each video frame increases the computation complexity in the SLAM pipeline. 
{The tracking latency observed in our system includes both inference time and updating time. Although the cumulative latency exceeds that required by the ORB algorithm, we can ensure the nearly immediate input of the tile-level importance results into the encoder module. This is crucial for minimizing the total time costs of client-edge collaboration, especially in complex traffic scenarios where each captured frame may contain a substantial number of feature points.}
Notably, the remarkable reduction in transmission latency makes our architecture outperform edge-SLAM in terms of E2E latency. Our architecture reduces the E2E latency by about $47\%$ in the real-measured network dataset and by $24.5\%$ in the simulated network dataset.

{
\emph{\textbf{Localization Accuracy:}}
We then conduct an experiment to evaluate the localization performance of two systems.
The pose estimation error determined by the camera pose estimation in ORB-SLAM2 is used to reflect the localization performance.
Fig. \ref{mappingerror_frame} presents the absolute pose error (APE) results for each frame while Fig. \ref{mappingerror_overall} shows the statistics.
The ground-truth results of APE are from the KITTI dataset \cite{geiger2013vision}.
 
Specifically, to obtain the APE results, we feed the video frames into both systems and then obtain the corresponding trajectory data that include the 3D camera pose information of each frame.
After that, we calculate the APE results based on an open-sourced tool \emph{evo} \cite{grupp2017evo}. 
From the statistics of Fig. \ref{mappingerror_overall}, it is observed that our system presents a more concentrated APE performance, which means that it can maintain a more stable and accurate localization performance compared to edge-SLAM. 

{
We use a subset from the KITTI dataset to compare the APE results obtained by our proposed architecture with those from the conventional edge SLAM. These results are obtained in the phase of system execution with a trained DRL-based configuration adaptation module.
The presentation of partial results if based on a condition that incorporating the ORB algorithm into the regional feature prediction module can ensure that the local module is applicable to all traffic scenarios, including those not previously encountered by the system. 
}}

\emph{\textbf{Evaluation of Prediction Model:}} To evaluate the prediction model, we employ the QoE performance obtained by the entire edge-assisted SLAM architecture and take the Optical Flow (OF) tracking as a baseline. OF tracking is used to track the apparent motion of image objects, which is a popular method in the field of computer vision. 
From Fig. \ref{qoekitti} it can be seen that our prediction model outperforms the OF tracking method in terms of estimating tile importance in SLAM. Further, Fig. \ref{qoekitti} also verifies the robustness of the configuration adaptation module in different network environments. 

\emph{\textbf{Convergence Performance of A3C:}} In this experiment, we evaluate the convergence performance of the A3C-based DRL algorithm in the configuration adaptation module. Specifically, we train the DRL agent using two baseline neural network architectures adopted from two well-established applications of the A3C algorithm, namely Pensieve\cite{mao2017neural} and OnRL\cite{zhang2020onrl}. The performance comparison in terms of convergence rates is shown in Fig. \ref{trainingconvergence}. 

Particularly, Pensieve is a popular Adaptive BitRate (ABR) selection system based on A3C, while OnRL is a recent application of A3C for live video applications. Both of them adopt the original learning principle of A3C from \cite{mnih2016asynchronous} but use different neural network architectures. Pensieve includes a combination of both a $1$D-CNN layer and an FC layer with $128$ elements, while OnRL contains only an FC layer with $64$ elements and an FC layer with $32$ elements. Based on the insights gained from Fig. \ref{trainingconvergence}, the optimal neural network architecture in our adopted A3C algorithm could be determined. 

   
 
\subsection{Scheduling Performance in Multi-Server Scenarios}
We first develop a faithful simulator to emulate the new edge-assisted SLAM architecture that operates in a wirelessly connected and multi-server scenario. The goal is to comprehensively evaluate the learning and scheduling performance of the scheduler.

\subsubsection{A Simulator for Task Scheduling in Multi-Server Scenarios}
The simulator is an extension of the Spark job simulator in \cite{mao2019learning} by modifying the Chameleon Cloud testbed\footnote{https://www.chameleoncloud.org}. 
{
The simulator is developed for training the scheduler offline. We are confident in the simulator because it effectively models the key components in the edge SLAM scenarios with multiple servers by capturing the following real-world effects: 
\begin{itemize}
\item[$\bullet$] It can access the profiling information of the SLAM pipeline, including the duration of each task or sub-task, and the dependencies between tasks in SLAM.
\item[$\bullet$] It can also access the processing time characteristics of SLAM tasks or sub-tasks by analyzing the sensory data. 
\item[$\bullet$] It simulates the network conditions from a client to multiple edge servers. We consider two kinds of network conditions to evaluate the robustness of the scheduler. 
\item[$\bullet$] It also simulates the state dynamics of each edge server, including both queuing and processing state.  
\end{itemize}

}

\textbf{Wireless Links:} In the simulator, we configure wireless environments among the scheduler, the user, and the edge servers. The goal of wireless configurations is to evaluate the robustness of the scheduler in different network conditions. Consequently, these configurations encompass two link settings between the user and each edge server. In the random link scenario, both the upload latency and throughput are generated from Gaussian distributions, whereas in the fixed link scenario, the latency and throughput are set as constant values. We conduct extensive experiments in both link settings, and the results will be illustrated in the subsequent subsections.   
  
\textbf{Tailored for Scheduling and Processing of SLAM Tasks:} We tailor the task scheduling and processing in the simulator by considering the property of SLAM tasks. Specifically, we first run the edge-assisted SLAM system in a single-server scenario with a well-trained configuration adaptation module by inputting multiple sequences from the KITTI dataset. 
A number of trajectories involving optimal Encoder/Decoder configurations for each frame in different link conditions could be 
obtained. From these trajectories, the scheduler treats each frame as a single task and thus can obtain the sizes of sub-tasks within each frame based on the optimal compression size of the frame. This can be accomplished by the SLAM pipeline.

In addition, the scheduler is capable of acquiring the status of wireless links and edge servers. In this regard, we employ synthetic trajectories while assuming the optimality of other components, enabling us to evaluate the scheduling performance in isolation. On the server side, we take the processing time of sub-tasks in the SLAM pipeline as the primary metric for the evaluation of scheduling performance. This is based on the premise that the improvement of localization and mapping primarily depends on the encoder/decoder and configuration adaptation components. Consequently, our main evaluation within the scheduler module revolves around the reduction of task completion time, which is quantified as the sum of the processing time of sub-tasks within each task.
  
In the simulator, there are a total of $10$ edge servers with diverse processing capabilities available for providing edge computing services. Although only a single user requests edge services, the sub-tasks within the SLAM pipeline exhibit diversity in terms of sizes, deadline requirements, and complexity since the trajectories are generated randomly and sufficiently. This ensures the generality of our experiments. 

The arrival rate of tasks is determined by the FPS of video frames from the KITTI dataset. To ensure a balanced workload across servers, we carefully configure the heterogeneous processing capabilities of each server. For simplicity, the processing capability of each server is fixed for different sub-tasks. 
  
\textbf{Incorporating User's Requirements:} We incorporate the requirements of the user into the simulator. The FPS of frames, specified in the KITTI dataset, determines the minimum processing rate of each task. That is, a processing rate lower than the FPS will lead to task accumulation in the server queue. 

For example, if FPS is set to $30$, the edge servers facilitated by scheduling must process a minimum of $30$ tasks per second in order to ensure seamless operation. FPS also limits the deadlines of sub-tasks within a task, which are taken as the user's requirements in the simulator. More specifically, with a given FPS, we set a fixed ratio between the duration of each 
type of sub-task and the task processing time required by the FPS. These ratios are then taken as the deadlines for these sub-tasks, representing the user's requirements. Importantly, the scheduler becomes aware of whether the processing time of a 
sub-task exceeds its deadline only after the completion of processing. In our experiments, the proposed input-dependent GP is used to capture the unknown and time-varying requirements by extracting the temporal correlation and then improving the training and scheduling performance of the scheduler. 

\begin{figure*}
\centering
  \begin{minipage}[t]{0.4\linewidth}
    \centering
    \includegraphics[scale=0.29]{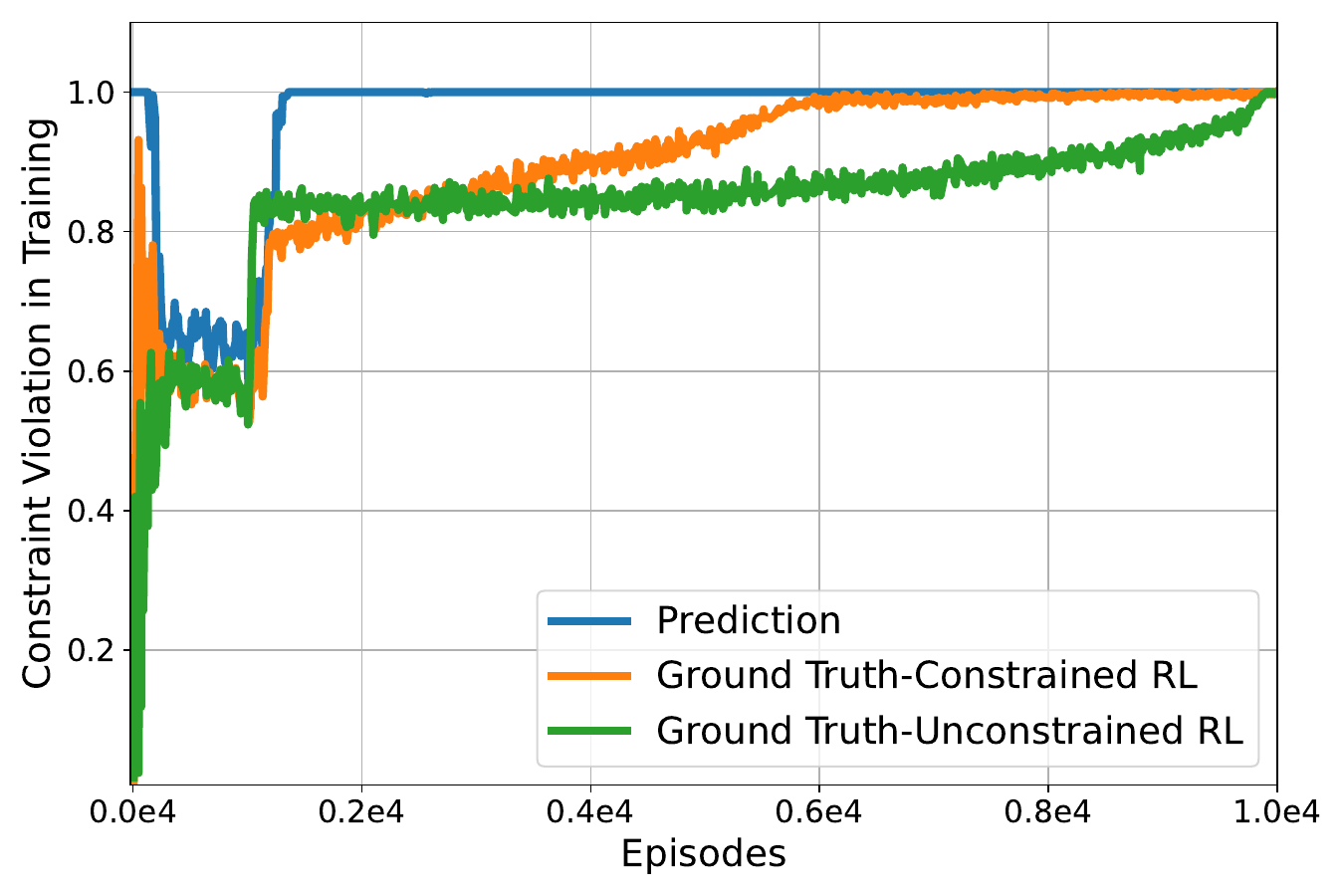}
    \caption{Constraint Violation in Fixed Links}
    \label{fixedchannels}
  \end{minipage}%
  \hspace{0.9cm}
  \begin{minipage}[t]{0.4\linewidth}
    \centering
    \includegraphics[scale=0.29]{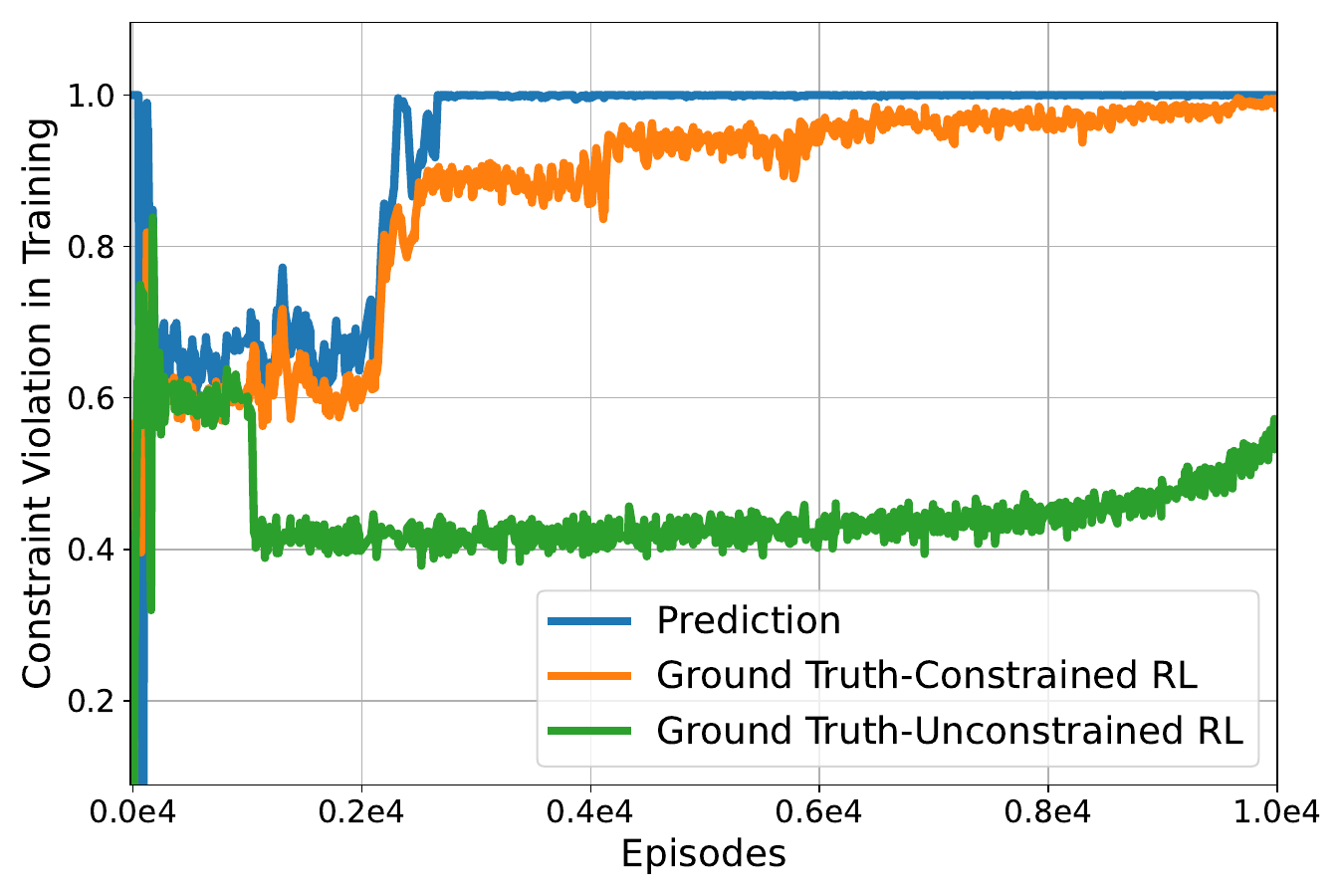}
    \caption{Constraint Violation in Random Links}
    \label{randomchannels} 
  \end{minipage}
  \vspace{-0.4cm}
\end{figure*}

\begin{figure}[!t]
\centering
\includegraphics[width=3.2in]{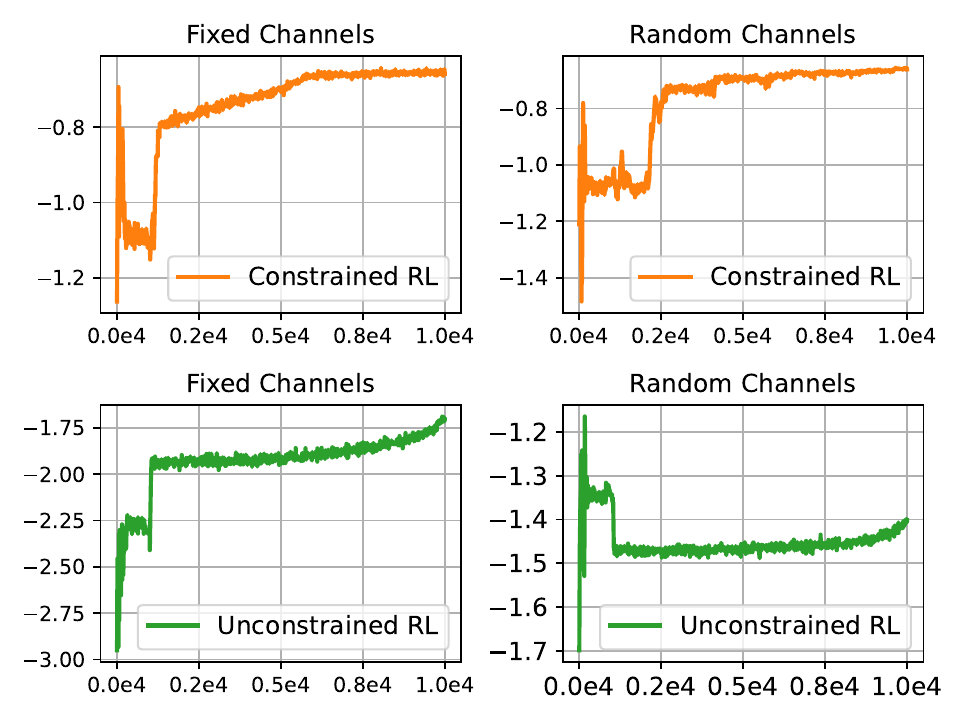}
\caption{Rewards in Constrained and Unconstrained RL}
\label{rewardswoconstrain}  
\vspace{-0.4cm}
\end{figure} 

\subsubsection{Experimental Setup}
We train the scheduler by incorporating Algorithm \ref{trainingscheduling} into the A3C architecture used in the configuration adaptation module. Specifically, we modify the network architecture in A3C to correspond to the simulator environments. 
The input layer corresponds to the state settings while the output layer will generate scheduling decisions. Both the $\gamma$ and the learning rate remain constant with the configuration adaptation module. Both ReLU function and Adam optimizer are adopted. We configure two hidden layers with $200$ and $128$ neurons for both the actor and critic networks.  
We employ the same parallel training architecture with $16$ parallel agents to improve the training efficiency. 
 
\subsubsection{{Baselines \& Results}}
Specifically, Fig. \ref{fixedchannels} and Fig. \ref{randomchannels} show the results of 
constraint violation during training, in order to verify the effectiveness {of constrained RL formulation and the input-dependent learning.}   
\emph{Ground Truth-Constrained DRL} adopts the constrained RL formulation in (\ref{constrainedRL}) with the reward function in (\ref{newreward}). On the other hand, \emph{Ground Truth-Unconstrained RL} adopts the traditional unconstrained RL framework with the following reward function (\ref{newreward_unconstrained}). In the results obtained from \emph{Ground Truth-Constrained RL} and \emph{Ground Truth-Unconstrained RL}, we employ the ground-truth information regarding the user's requirements, which enables us to place a strong emphasis on the enhanced learning capability of the constrained RL formulation. The results labeled in \emph{Prediction} are the one-step prediction of state safety obtained by the fitted input-dependent GP model and used to indicate constraint violation. An unsafe state refers to a state in which the constraint specified in (\ref{constrainedRL}) will be violated.
\begin{equation}
\begin{aligned}
\centering
\label{newreward_unconstrained}
r_t(\dot{a}_{t},\dot{s}_{t},\hat{\dot{s}}_{t+1})=r_t(\dot{a}_{t},\dot{s}_{t})+\mathcal{P}
\end{aligned}
\end{equation}

In the subsequent experiments, the results labeled as \emph{Constrained RL} are obtained by executing Algorithm \ref{trainingscheduling}. On the other hand, \emph{Unconstrained RL} refers to executing Algorithm \ref{trainingscheduling} by neglecting lines $13-21$. 

\begin{figure*}
\centering
\hspace{-0.9cm}
  \begin{minipage}[t]{0.32\linewidth}
    \includegraphics[scale=0.27]{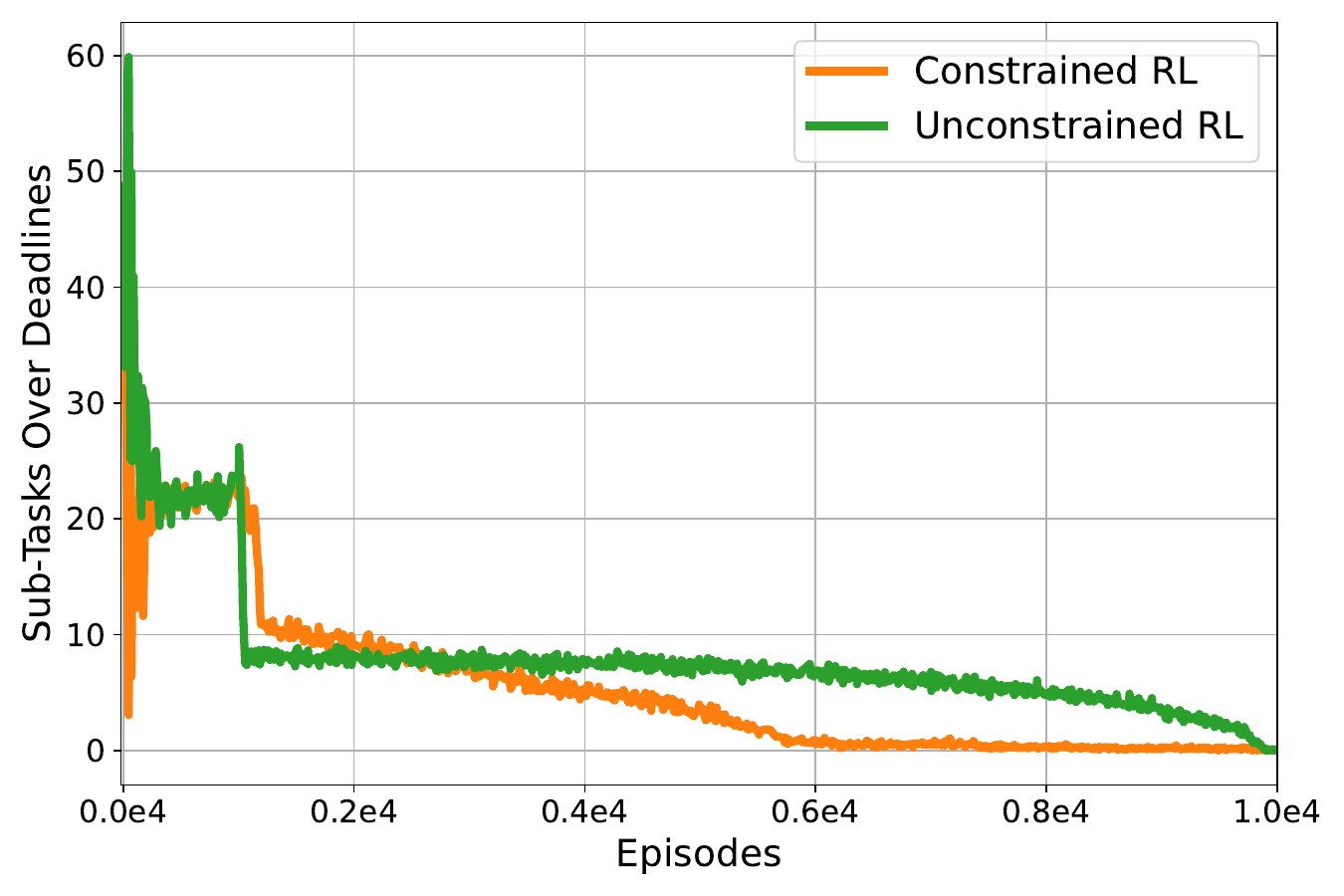}
    \caption{Sub-Tasks Over Deadlines Under Fixed Links}
    \label{fixedchannels_tasks_train}
  \end{minipage}%
  \hspace{0.13cm}
  \begin{minipage}[t]{0.32\linewidth}
    \centering
    \includegraphics[scale=0.27]{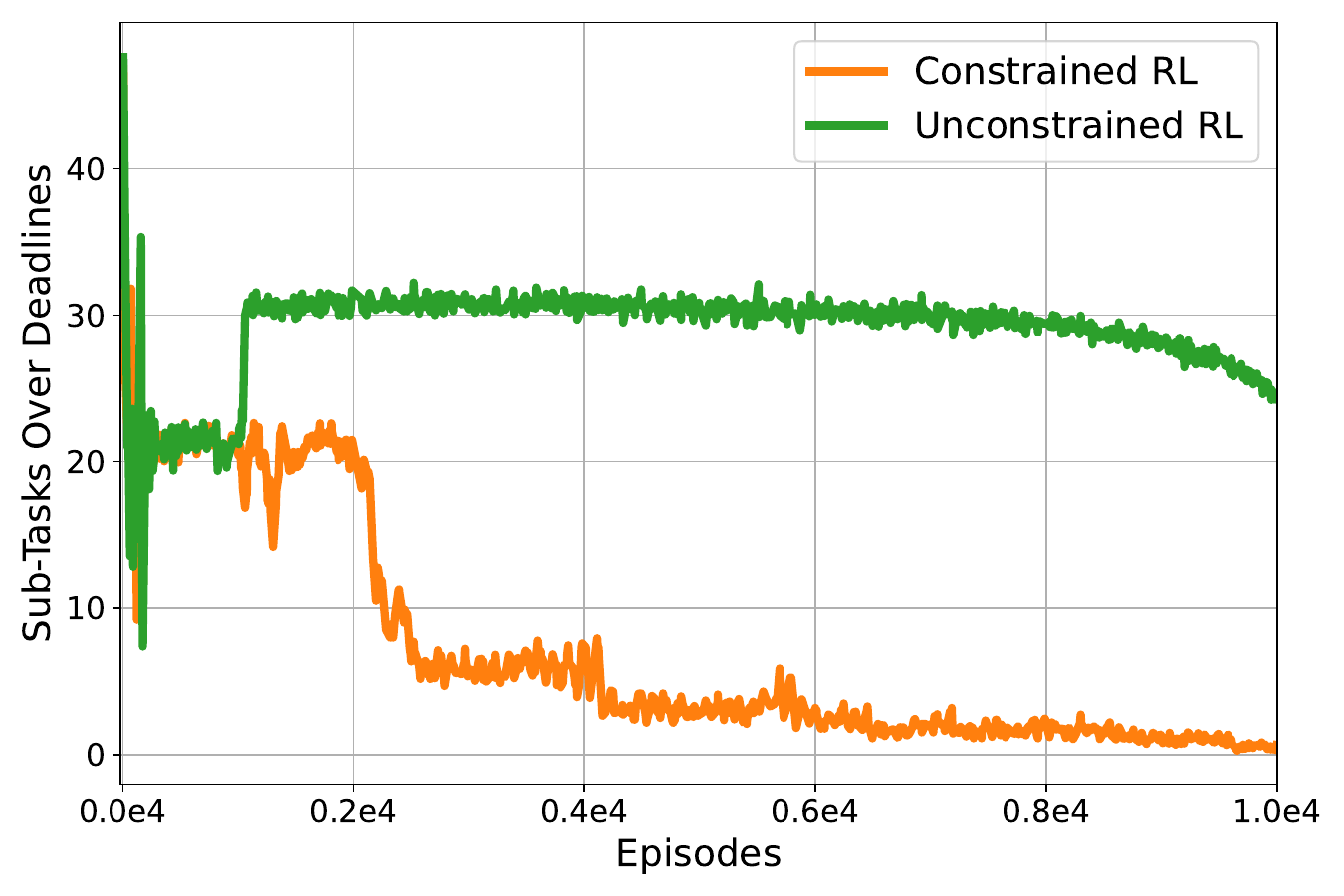}
    \caption{Sub-Tasks Over Deadlines Under Random Links}
   \label{randomchannels_tasks_train} 
  \end{minipage}
  \hspace{0.13cm}
    \begin{minipage}[t]{0.32\linewidth}
    \centering
    \includegraphics[scale=0.27]{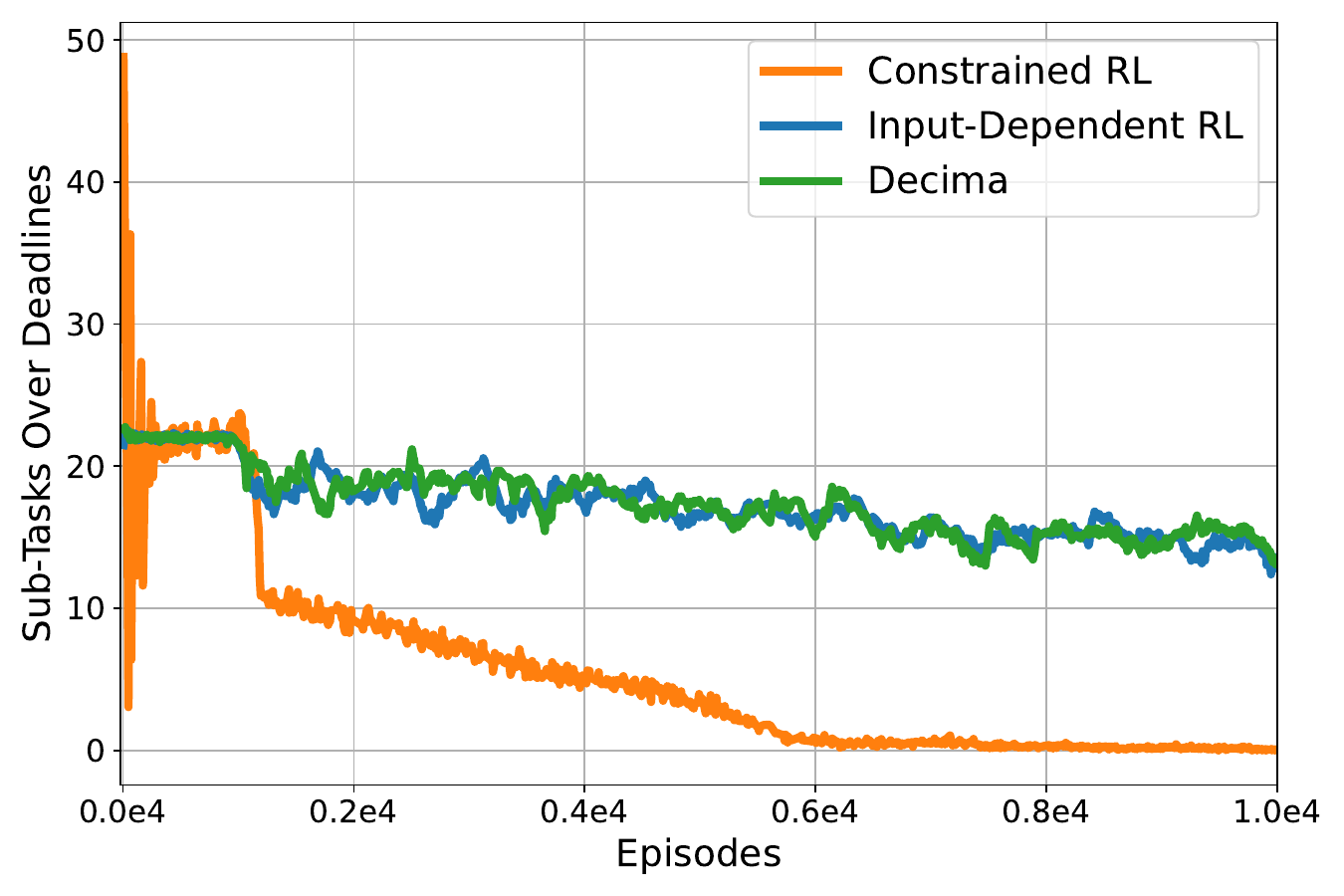}
    \caption{Sub-Tasks in Training Under Fixed Links}
 \label{comparison_fixedchannels_tasks_train}
  \end{minipage}%
  \hspace{-0.4cm}
\end{figure*}

Fig. \ref{rewardswoconstrain} to Fig. \ref{randomchannels_tasks_train} further verify the advantages of the input-dependent GP model by comparing \emph{Constrained RL} and \emph{Unconstrained RL} under different link conditions. 

\begin{figure*}
\centering
\hspace{-0.9cm}
  \begin{minipage}[t]{0.32\linewidth}
    \centering
    \includegraphics[scale=0.27]{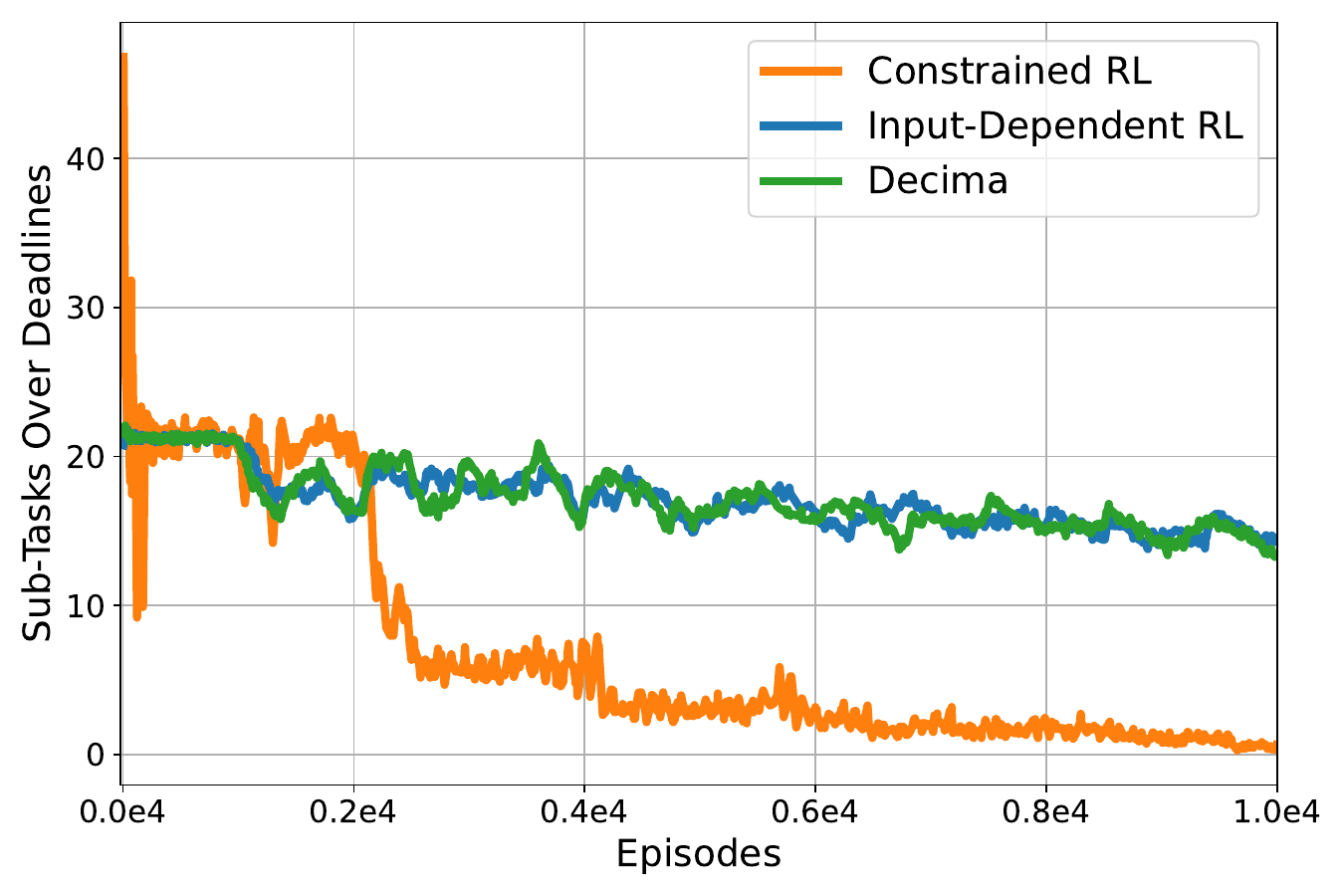}
    \caption{Sub-Tasks in Training Under Random Links}
    \label{comparison_randomchannels_tasks_train} 
  \end{minipage}
  \hspace{0.07cm}
  \begin{minipage}[t]{0.32\linewidth}
    \centering
    \includegraphics[scale=0.27]{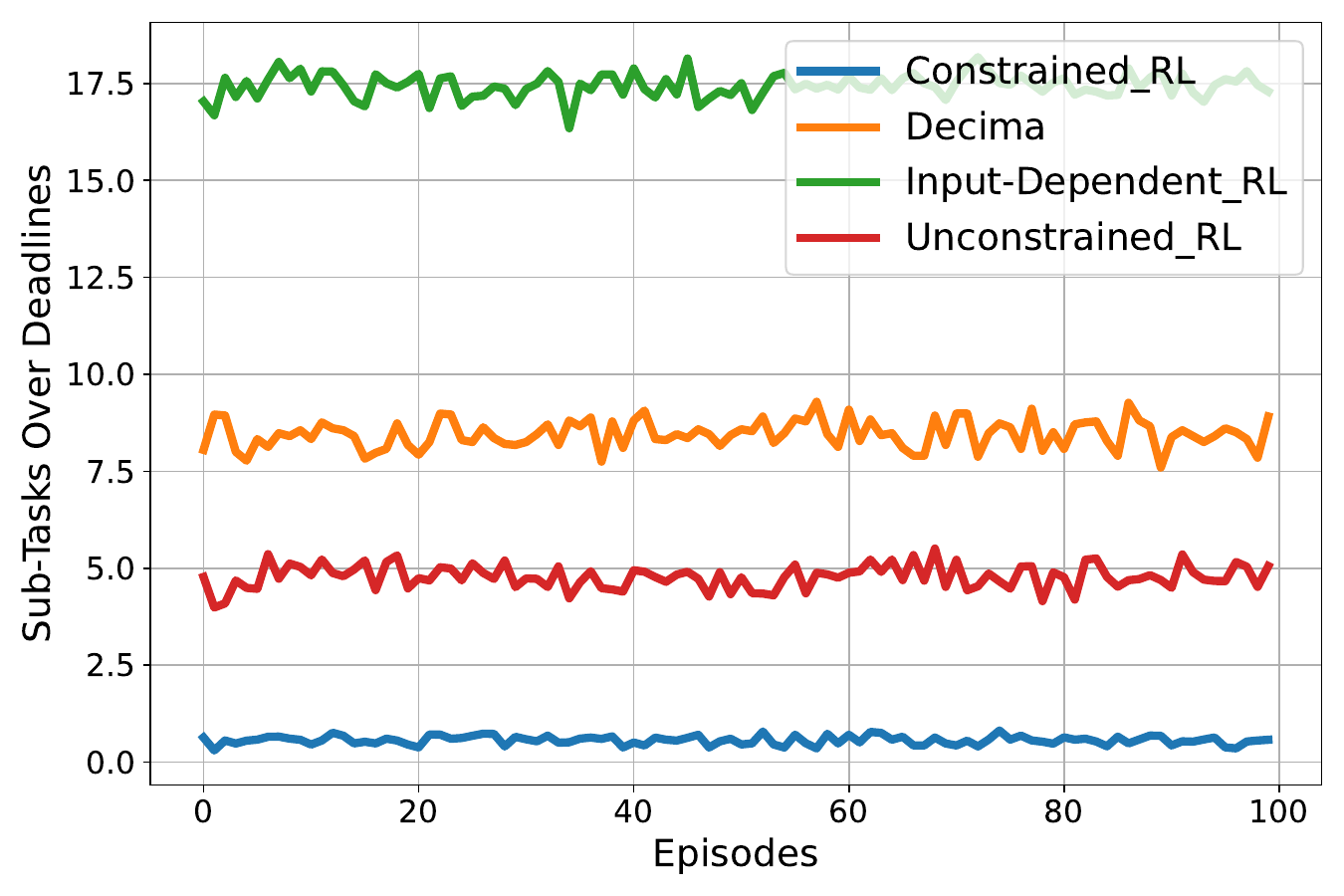}
    \caption{Sub-Tasks in Testing Under Fixed Links}
    \label{comparison_fixedchannels_tasks_test}
  \end{minipage}%
  \hspace{0.16cm}
  \begin{minipage}[t]{0.32\linewidth}
    \centering
    \includegraphics[scale=0.27]{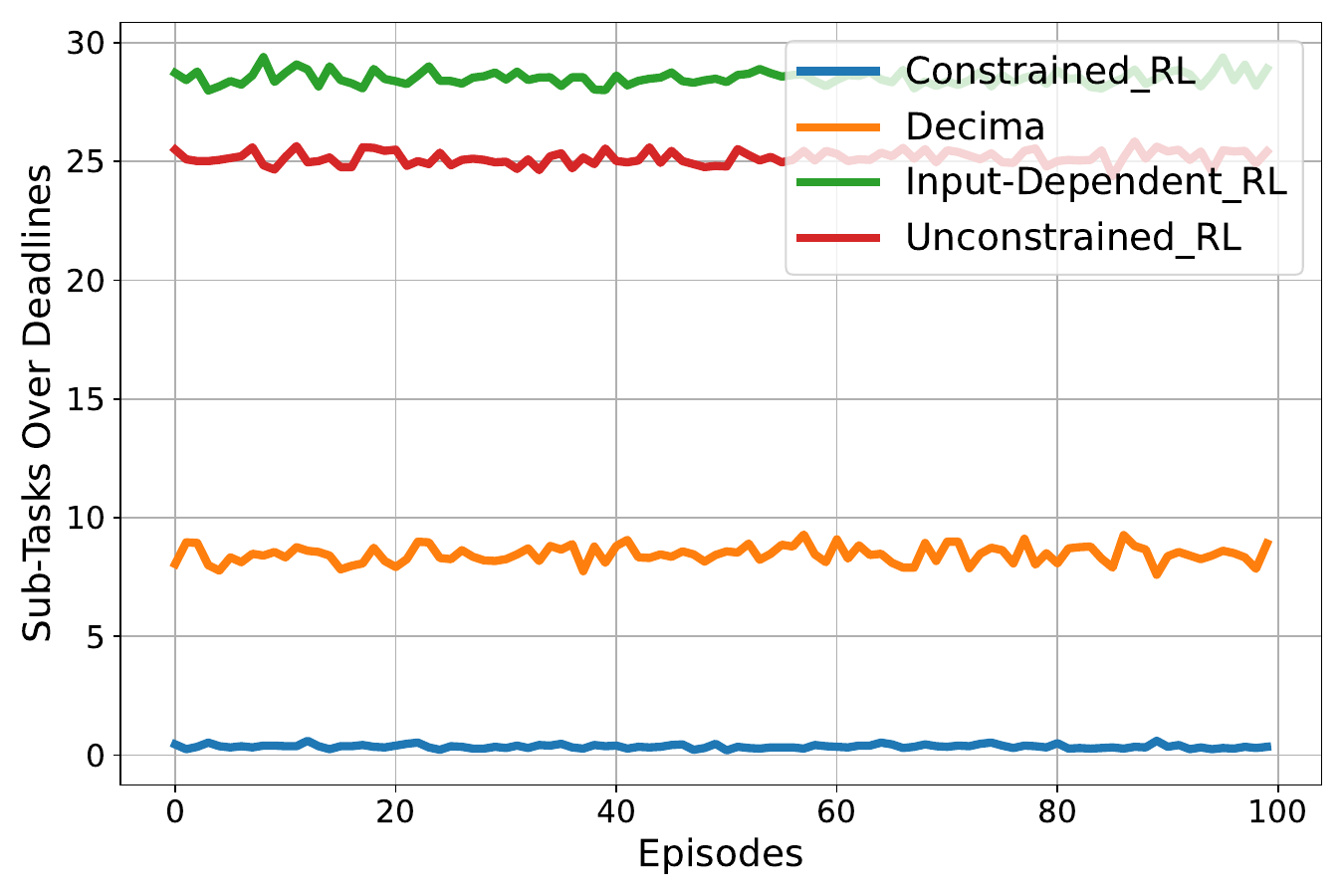}
    \caption{Sub-Tasks in Testing Under Random Links}
    \label{comparison_randomchannels_tasks_test} 
  \end{minipage}
  \hspace{-0.4cm}
  \vspace{-0.4cm}
\end{figure*}

In addition to the \emph{Constrained RL} with Algorithm \ref{trainingscheduling}, we compare two baseline scheduling methods, namely \emph{Input-Dependent RL}\cite{mao2019variance} and \emph{Decima}\cite{mao2019learning}. 
\emph{Input-Dependent RL} is an RL-based scheduling algorithm that improves the A2C algorithm by incorporating input-dependent baselines instead of state-dependent baselines, resulting in improved variance reduction.  
\emph{Decima} develops an embedding method that employs Graph Neural Networks (GNN) within the RL framework to capture the dependency 
between sub-tasks when making scheduling decisions. 
We implement both \emph{Input-Dependent RL} and \emph{Decima} in the simulator and set the same reward function with \emph{Unconstrained RL}.
The training performance is depicted in Fig. \ref{comparison_fixedchannels_tasks_train} and 
Fig. \ref{comparison_randomchannels_tasks_train} while the testing performance is presented in Fig. \ref{comparison_fixedchannels_tasks_test} and Fig. \ref{comparison_randomchannels_tasks_test}.

\subsection{Results}
\emph{\textbf{Evaluation of Constrained RL Formulation and Input-Dependent Learning:}}
In Fig.~\ref{fixedchannels} and Fig.~\ref{randomchannels}, the results of constraint violation indicators are plotted. We set the indicator as $1$ if the constraint is satisfied at each step, and $0$ otherwise. The mean value of the indicators from each episode is then plotted. 
First, from the curve obtained by \emph{Ground Truth-Constrained RL}, the trained scheduler guarantees that the constraint is satisfied gradually, which means that the scheduling decisions can satisfy the user's requirements with a high probability.
In contrast, \emph{Ground Truth-Unconstrained RL} exhibits a limited ability to prevent constraint violation, particularly in the scenario of random links. It can be confirmed that the constrained RL formulation achieves an enhanced learning ability in the input-driven system. Second, it can be seen that \emph{Prediction} has a similar trend with \emph{Ground Truth-Constrained RL}, which validates that the well-fitted input-dependent GP model effectively maintains a high level of prediction accuracy for state safety.
From the comparison of Fig.~\ref{fixedchannels} and Fig.~\ref{randomchannels}, it can be seen that the constrained RL formulation is robust to different network conditions.
  
\emph{\textbf{Training Performance:}} The training performance of \emph{Constrained RL} and \emph{Unconstrained RL} is shown in Fig.~\ref{rewardswoconstrain} to Fig.~\ref{randomchannels_tasks_train}. Fig.~\ref{rewardswoconstrain} showcases the rewards of two methods trained in both link scenarios. The results demonstrate that \emph{Constrained RL} is able to achieve a faster reward convergence in comparison to \emph{Unconstrained RL}, which thus verifies its effectiveness and robustness. It is worth noting that the results are presented separately due to the utilization of different reward functions for \emph{Constrained RL} and \emph{Unconstrained RL}.

We collect the number of sub-tasks that exceed their deadlines during training and plot them in Fig.~\ref{fixedchannels_tasks_train} and Fig.~\ref{randomchannels_tasks_train}, in order to evaluate the ability of \emph{Constrained RL} in meeting the user's requirements. Notably, the user's requirements are set as unknown to the system in advance. In both link scenarios, \emph{Constrained RL} consistently demonstrates a significant ability to ensure the timely completion of all sub-tasks. In contrast, \emph{Unconstrained RL} fails to achieve that in the scenario of random links. We can confidently conclude that the scheduler utilizing Algorithm \ref{trainingscheduling} can effectively schedule tasks without violating users' requirements, even in cases where the requirements are known in advance.



\emph{\textbf{Comparison of Training Performance:}}  
Fig. \ref{comparison_fixedchannels_tasks_train} and Fig. \ref{comparison_randomchannels_tasks_train} further present the results by comparing \emph{Constrained RL} and two baseline scheduling algorithms in both link scenarios. Both \emph{Input-Dependent RL} and \emph{Decima} are trained using the same experimental settings as \emph{Unconstrained RL} by taking (\ref{newreward_unconstrained}) as their reward function. 
 
In both link scenarios, \emph{Constrained RL} outperforms both \emph{Input-Dependent RL} and \emph{Decima}.
Particularly, \emph{Input-Dependent RL} emphasizes variance reduction by introducing a novel input-dependent baseline within the traditional RL framework, whereas \emph{Decima} utilizes a feature embedding method based on GNN. 
Although \emph{Input-Dependent RL} reduces the variance of RL-enabled scheduling, the dynamic nature of wireless connections introduces additional complexity to the input-driven system, leading to a deterioration in its performance.
Additionally, both \emph{Input-Dependent RL} and \emph{Decima} do not explicitly account for safety in state transitions, leading to inferior performance in meeting the system's constraints imposed by users' requirements.

\emph{\textbf{Comparison of Testing Performance}}
Fig.~\ref{comparison_fixedchannels_tasks_test} and Fig.~\ref{comparison_randomchannels_tasks_test} present the testing results by comparing \emph{Constrained RL} with three baselines in different network scenarios. To ensure a faithful and precise comparison, we adopt two approaches in our experiments. First, the testing results are obtained by executing online inference with the trained agents on previously unseen input sequences and setting the test episode to $100$. Second, during online inference, we exclude the GP-related operations specific to \emph{Cosntrained RL}. That is, during the training process, the state transition is rectified by the input-dependent GP model, while during online inference on unseen sequences, the state transition of all methods is not intervened.

From Fig.~\ref{comparison_fixedchannels_tasks_test} and Fig.~\ref{comparison_randomchannels_tasks_test}, it can be seen that the proposed \emph{Constrained RL} achieves optimal performance indicated by the results consistently close to zero. In other words, it ensures that the user's requirements can be satisfied in most episodes. In addition, \emph{Constrained RL} demonstrates better stability compared to other baselines throughout the testing experiments. In the scenario of fixed links, \emph{Input-Dependent RL} obtains the worst performance, indicated by the results consistently near $17.5$. In the scenario of random links, the case becomes worse as the results obtained by \emph{Input-Dependent RL} are about $28$. Another conclusion we can see in the scenario of random links is that \emph{Decima} outperforms \emph{Unconstrained RL}, which is 
contrary to that in the scenario of fixed links. Both \emph{Constrained RL} and \emph{Decima} obtain similar performance in both scenarios, while the performance of the other two baselines deteriorates under the random links. This demonstrates the robustness of \emph{Constrained RL} and \emph{Decima}. Note that, the results in both Fig.~\ref{comparison_fixedchannels_tasks_test} and Fig.~\ref{comparison_randomchannels_tasks_test} differ from the converged results in the training experiments, which is because we set different random seeds. 

\section{Related Works}
\label{relatedwork} 

\subsection{SLAM \& Edge SLAM}
Recently, SLAM has become a critical enabling technology for many emerging human-centered applications, such as AR/VR \cite{savva2019habitat}, HD map \cite{ahmad2020carmap, qiu2018avr}, and autonomous navigation \cite{vithalani2020autonomous}. 
In terms of mobile robotics systems with limited computation resources, edge-assisted SLAM has drawn substantial attention since it has a great potential to reduce the computation burden on the client side, such as autonomous vehicles \cite{zhang2019mobile}, UVAs \cite{chen2021edge}. The existing works that are closely related to edge SLAM include
\cite{braud2020multipath}\cite{ahmad2020carmap}\cite{xu2020edge}\cite{ben2020edge}\cite{liu2021edgesharing}\cite{xu2022swarmmap}. 

Specifically, Tristan et al. \cite{braud2020multipath} and Jingao et al. \cite{xu2020edge} focus on offloading sub-tasks of mobile SLAM, i.e., the SLAM system on handheld devices. Tristan et al. \cite{braud2020multipath} develop a scheduler to dispatch sub-tasks over two edge servers while Jingao et al. \cite{xu2020edge} consider the collaboration of two nearby handheld devices for edge SLAM. Fawad et al. \cite{ahmad2020carmap} propose a light HD map construction system by formalizing an edge-SLAM architecture for autonomous vehicles. The authors verify that the compressed vision data is an efficient way in edge SLAM. Edge-SLAM \cite{ben2020edge} is the first work of system-level decomposition for edge SLAM, i.e., the authors propose to decompose sub-tasks and offload part of them into the edge server. Luyang et al. \cite{liu2021edgesharing} construct a crowd SLAM at the edge by collecting data from clients and then sharing global SLAM information with them. Jingao et al. \cite{xu2022swarmmap} design a framework for collaborative SLAM at the edge, in order to scale up collaborative SLAM services. Different from existing SLAM-related works, this work focuses more on the adaptation of edge SLAM, i.e., adaptive offloading and adaptive scheduling, especially in the multi-server scenario. In addition, we highlight the importance of user requirements in SLAM systems and propose an efficient learning framework to address that.
  
\subsection{Learning to Offloading} 
Our work gains important insights from the existing works that investigate the offloading strategies for vision-supported applications e.g., AR/VR, video analytics \cite{guo2018small, galanopoulosautoml, wang2020joint, ren2021adaptive}. Specifically, an important principle in those works is to strike the balance between inference accuracy and communication overhead via offloading compressed vision data. However, the compression options are usually configured for the whole frame. The diverse sensitivity of different frame regions to data distortion caused by compression has not been considered. For example, the frame regions containing small objects need a higher resolution to guarantee object detection accuracy according to the conclusion in \cite{guo2018small}. Notably, this problem will become severe in vSLAM since pixel-level feature extraction in vSLAM (e.g., the ORB algorithm \cite{mur2017orb}) is more sensitive to data distortion than object detection. 

Several works focus on tile-oriented compression solutions \cite{liu2019edge, wang2021edgeduet, du2020server}. Similarly, we further investigate the tile-level importance on the client side and configure adaptive compression in various network environments.
 
\subsection{Learning to Scheduling} 
Scheduling tasks among multiple servers or between the server and Cloud is an important research topic in the computer community. Several recent works provide us with valuable insights \cite{mao2019learning}\cite{zhang2020towards}\cite{ayala2019vrain}\cite{shen2023collaborative}. Specifically, Hongzi et al. \cite{mao2019learning} proposes a learning-based scheduling algorithm to efficiently schedule jobs whose tasks have complex dependencies. The authors employ GNN to extract the dependency property of tasks. Shigeng et al. \cite{zhang2020towards} focus on the problem of partition and offloading of DNN models and propose a Directed Acyclic Graph (DAG)-based problem formulation with an optimal partition algorithm. Jose et al. \cite{ayala2019vrain} propose an autoencoder-based mining method to extract latency representation from high-dimensional context data for resource allocation in the virtualization of radio access networks. The high-dimensional context data in \cite{ayala2019vrain} is heterogeneous, e.g., the combination of traffic and signal quality patterns. The difference is that our work focuses on the vision data. Shihao et al. \cite{shen2023collaborative} develop a scheduling framework for dispatching tasks among the edge and Cloud in the kubernetes system. 

Nevertheless, most of the existing scheduling-related works partially focus on load balancing or task processing performance. In our work, users' requirements (e.g., tasks processing DDL) highly affect the SLAM performance while those requirements are usually implicit and time-varying. 

{
\subsection{Other Vision or Non-Vision Based Methods} 
One of the advantages of visual SLAM is its ability to simultaneously construct environmental map and estimate the camera's pose, which facilitate the localization, navigation, and control of mobile robotic systems. 


In recent years, many new technologies have been exploited to enhance 
devices' ability to understand their environment, thereby improving the mapping and localization performance. In the field of computer vision, 3D scene reconstruction is an emerging technology that enhances the understanding of both indoor and outdoor environments. 
Recently, 3D scene reconstruction has been significantly improved with the use of generative models, such as the diffusion model-based SceneDiffuser \cite{huang2023diffusion} and the Gaussian splitting-based Drivinggaussian \cite{zhou2024drivinggaussian}. 
Another promising approach in computer vision for understanding 3D environments is based on analyzing geometric points. 
For example, point clouds are an important type of geometric point set. The problem of how to extract useful environment-related features by processing point clouds has been widely studied via deep learning methods\cite{qi2017pointnet++}. 

In addition to vision-based methods, there are also many promising technologies that leverage wireless signals to understand environments. Radio-based SLAM \cite{leitinger2019belief} is a type of SLAM mechanism that constructs radio maps and localizes devices by estimating channel multipath angles. 
Another efficient approach to understanding environments is based on the Neural Radiance Fields \cite{zhao2023nerf2}. The key idea is to represent a scene or object as a neural radiance field, which is trained using a small number of signal measurements. By efficiently synthesizing the spatial spectrum given the position of a transmitter (TX) or receiver (RX), It has demonstrated excellent performance in localization and wireless communications.

Although these aforementioned methods can achieve decent mapping or localization performance in their targeted scenarios, it remains unclear whether they can be effectively applied in autonomous driving scenarios. Most existing 3D scene reconstruction methods require extensive training. The performance of radio-based methods are easily affected by wireless interference, and will be degraded in complex propagation environments. }

\section{Conclusion} \label{conclusion}
This paper develops an innovative edge-assisted SLAM architecture for the general adoption of SLAM on mobile robotic systems. With the new architecture, the SLAM system can largely benefit from the abundant computational resources at edge servers with reduced communication overhead, improved SLAM performance, satisfied users' requirements, and a balanced workload of servers. Particularly, the significance of the importance-aware local data processing is obvious because of its valuable contribution to client-edge collaboration. This approach effectively reduces redundancy while highlighting essential content within each frame. Configuration adaptation is an inevitable component as it enables the system to be robust in various environmental dynamics. The input-dependent learning framework provides an important insight into how to learn the temporal correlation from the implicit and time-varying requirements.



\bibliographystyle{IEEEtran}
\bibliography{sample-base}

\begin{IEEEbiography}[{\includegraphics[width=1.4in,height=1.25in,clip,keepaspectratio]{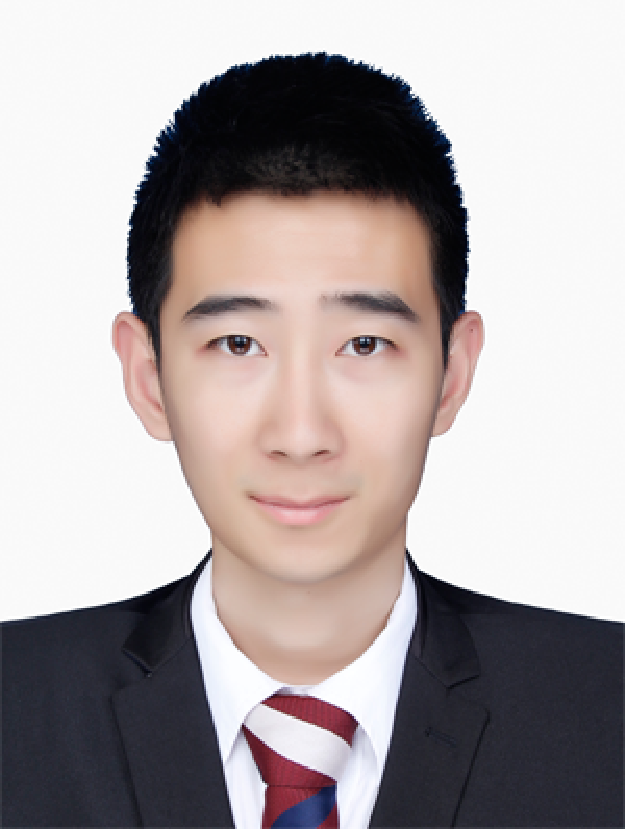}}]
{Yao Zhang} received his Ph.D. degree in communication and information system from Xidian University, Xi'an, China, in 2020. He served as a Research Assistant and Post-Doctoral Fellow at The Hong Kong Polytechnic University in 2019 and 2021, respectively. From 2021 to 2023, he was a Post-Doctoral Researcher at Northwestern Polytechnical University, Xi'an, China, where he is currently an Associate Professor. His research interests include CrowdNet, human-AI collaboration, mobile edge computing, edge AI, and networked autonomous driving.
\end{IEEEbiography}
\vspace{-1cm}

\begin{IEEEbiography}[{\includegraphics[width=1.4in,height=1.25in,clip,keepaspectratio]{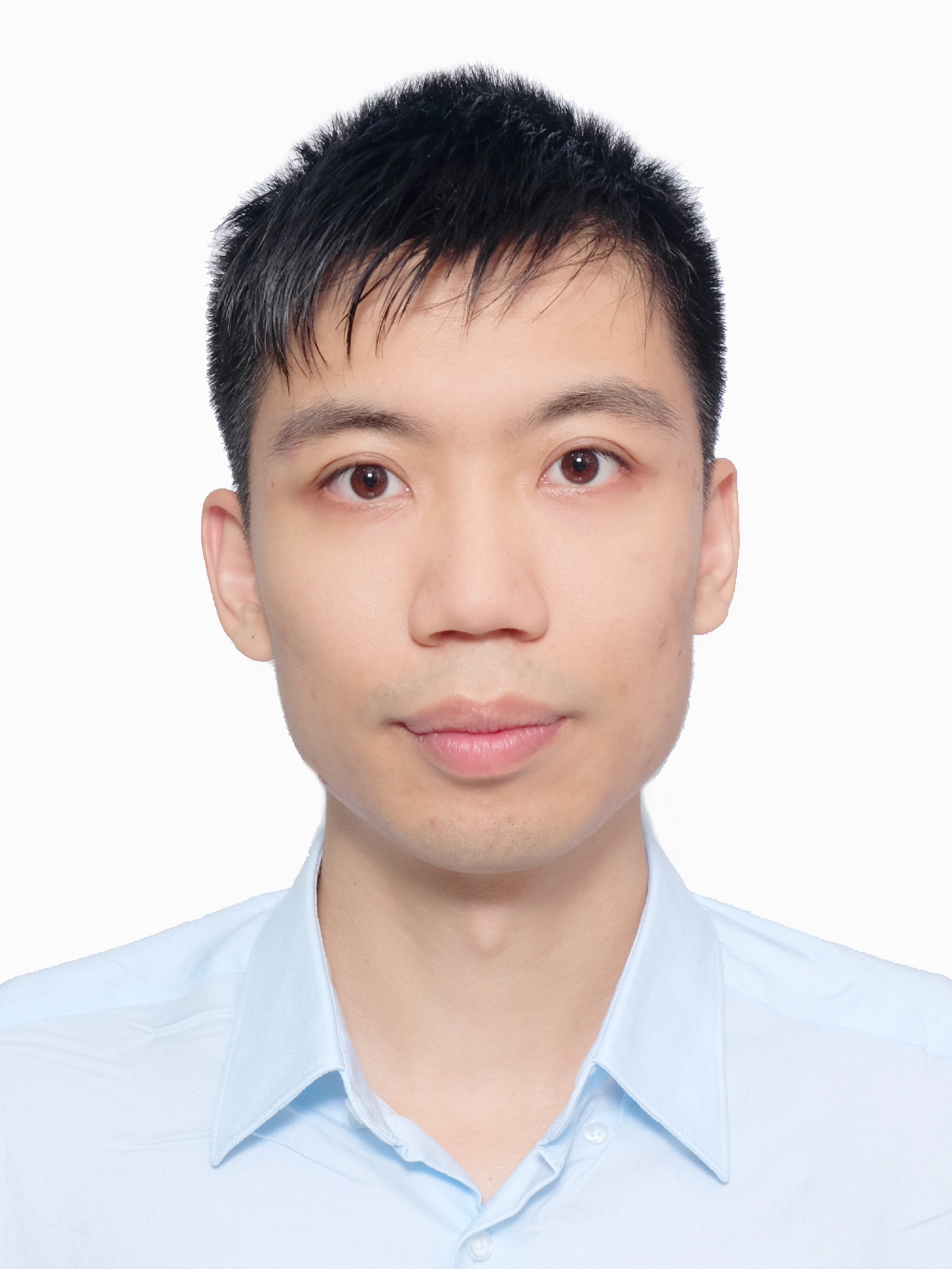}}]{Yuyi Mao} received the B.Eng. degree in Information and Communication Engineering from Zhejiang University (ZJU), Hangzhou, China, in 2013, and the Ph.D. degree in Electronic and Computer Engineering from The Hong Kong University of Science and Technology (HKUST), Hong Kong, in 2017. He was a Lead Engineer with The Hong Kong Applied Science and Technology Research Institute Co., Ltd. (ASTRI), Hong Kong, a Senior Researcher with the Theory Lab, 2012 Labs, Huawei Tech. Investment Co., Ltd., Hong Kong, and a Research Assistant Professor with the Department of Electrical and Electronic Engineering, The Hong Kong Polytechnic University (PolyU), Hong Kong. He is currently an Assistant Professor with the School of Computer Science and Engineering, Macau University of Science and Technology (MUST), Macau. His research interests include wireless communications and networking, mobile edge computing and learning, and wireless artificial intelligence.

Dr. Mao was the recipient of the 2021 IEEE Communications Society Best Survey Paper Award and the 2019 IEEE Communications Society $\&$ Information Theory Society Joint Paper Award. He was also recognized as an Exemplary Reviewer of the IEEE WIRELESS COMMUNICATIONS LETTERS in 2019 and 2021, and the IEEE TRANSACTIONS ON COMMUNICATIONS in 2020. He is an Editor of the IEEE WIRELESS COMMUNICATIONS LETTERS, and an Associate Editor of the EURASIP JOURNAL ON WIRELESS COMMUNICATIONS AND NETWORKING and the HKIE TRANSACTIONS.
\end{IEEEbiography}
\vspace{-1cm}

\begin{IEEEbiography}[{\includegraphics[width=1.4in,height=1.25in,clip,keepaspectratio]{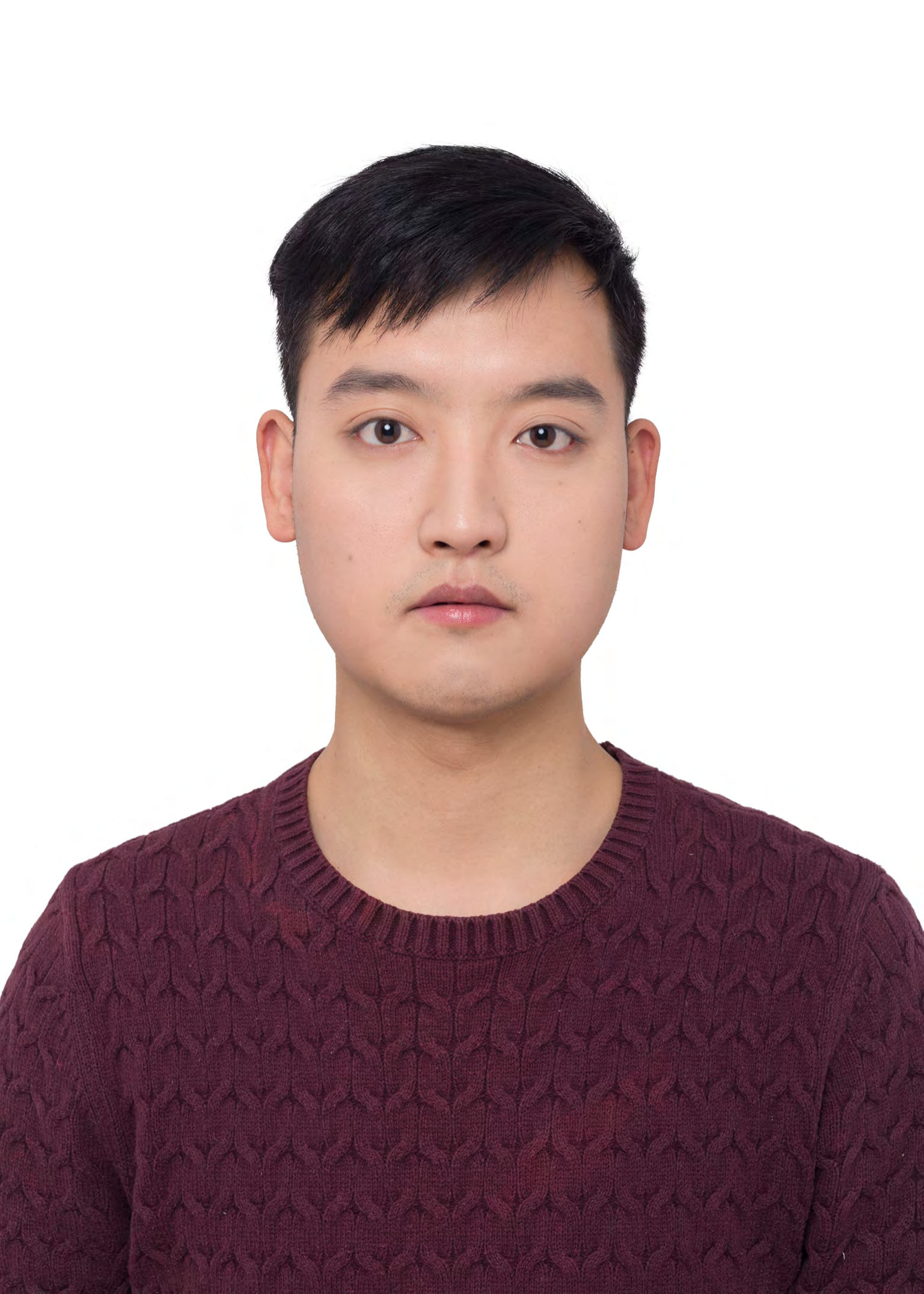}}]{Hui Wang} received the M.S. degree in communications and information system from Xidian University, Xi'an, China, in 2020. He is currently pursuing the Ph.D. degree with the School of Computer Science, Northwestern Polytechnical University, Xi'an, China. His research interests include ubiquitous computing, human-machine computing, and artificial intelligence.
\end{IEEEbiography}
\vspace{-1cm}

\begin{IEEEbiography}[{\includegraphics[width=1.4in,height=1.25in,clip,keepaspectratio]{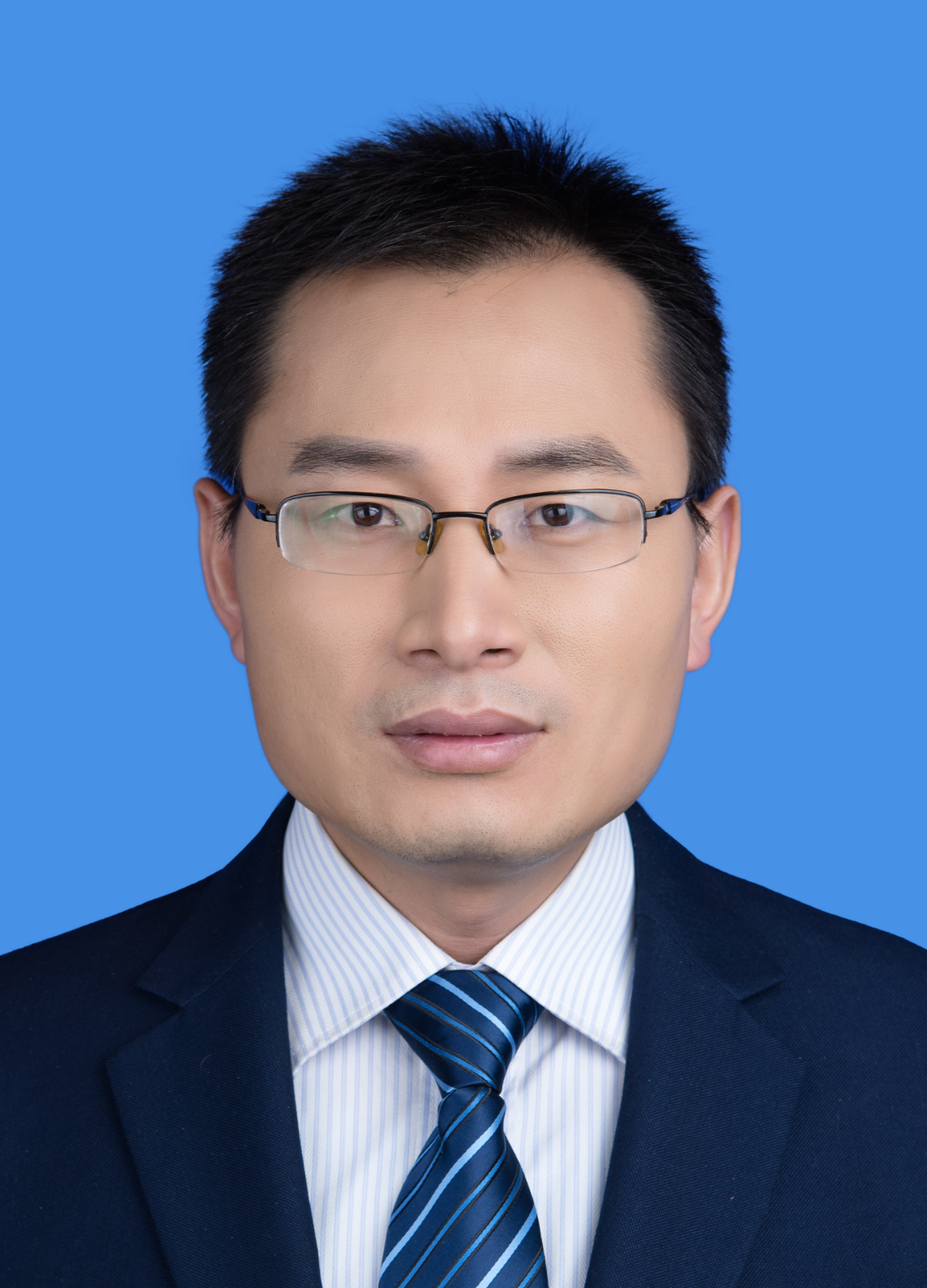}}]{Zhiwen Yu} received the Ph.D. degree of engineering in computer science and technology from Northwestern Polytechnical University, Xi'an, China, in 2005. He is currently with Harbin Engineering University, Harbin, Heilongjiang, China, and also a professor at Northwestern Polytechnical University, Xi'an, China. He has worked as a research fellow with the Academic Center for Computing and Media Studies, Kyoto University, Japan, from February 2007 to January 2009, and a post-doctoral researcher with the Information Technology Center, Nagoya University, Japan, in 2006-2007. His research interests include pervasive computing, context-aware systems, human-computer interaction, mobile social networks, and personalization. He is the associate editor or editorial board member for the IEEE Communications Magazine, the IEEE Transactions on Human-Machine Systems, Personal and Ubiquitous Computing, and Entertainment Computing (Elsevier). He was the general chair of IEEE CPSCom'15, and IEEE UIC'14. He served as a vice program chair of PerCom'15, the program chair of UIC'13, and the workshop chair of UbiComp'11. He is a senior member of the IEEE, a member of the ACM, and a council member of the China Computer Federation (CCF).
\end{IEEEbiography}

\begin{IEEEbiography}
[{\includegraphics[width=1.3in,height=1.25in,clip,keepaspectratio]{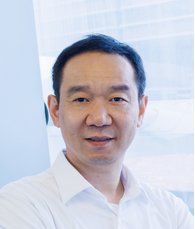}}]
{Song Guo} is a Full Professor in the Department of Computer Science and Engineering at Hong Kong University of Science and Technology. He also holds a Changjiang Chair Professorship awarded by the Ministry of Education of China. His research interests are mainly in the areas of big data, edge AI, mobile computing, and distributed systems. With many impactful papers published in top venues in these areas, he has been recognized as a Highly Cited Researcher (Web of Science) and received over 12 Best Paper Awards from IEEE/ACM conferences, journals and technical committees. Prof. Guo is the Editor-in-Chief of IEEE Open Journal of the Computer Society. He has served on IEEE Communications Society Board of Governors, IEEE Computer Society Fellow Evaluation Committee. He has also served as chair of organizing and technical committees of many international conferences. Prof. Guo is an IEEE Fellow and an ACM Distinguished Member.
\end{IEEEbiography}
\vspace{-2cm}

\begin{IEEEbiography}
[{\includegraphics[width=1.4in,height=1.25in,clip,keepaspectratio]{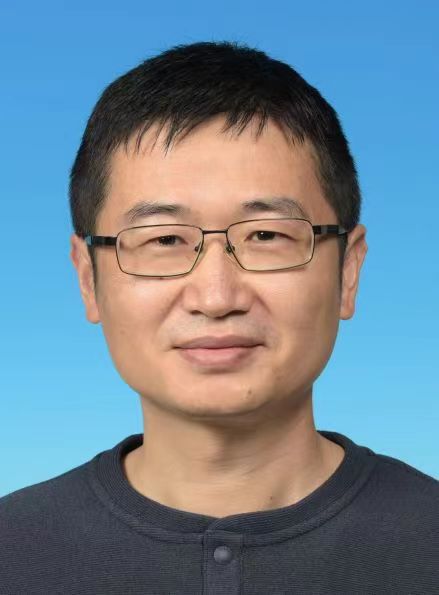}}]{Jun Zhang} received the B.Eng. degree in Electronic Engineering from the University of Science and Technology of China in 2004, the M.Phil. degree in Information Engineering from the Chinese University of Hong Kong in 2006, and the Ph.D. degree in Electrical and Computer Engineering from the University of Texas at Austin in 2009. He is an Associate Professor in the Department of Electronic and Computer Engineering at the Hong Kong University of Science and Technology. His research interests include wireless communications and networking, mobile edge computing and edge AI, and cooperative AI. Dr. Zhang co-authored the book Fundamentals of LTE (Prentice-Hall, 2010). He is a co-recipient of several best paper awards, including the 2021 Best Survey Paper Award of the IEEE Communications Society, the 2019 IEEE Communications Society \& Information Theory Society Joint Paper Award, and the 2016 Marconi Prize Paper Award in Wireless Communications. Two papers he co-authored received the Young Author Best Paper Award of the IEEE Signal Processing Society in 2016 and 2018, respectively. He also received the 2016 IEEE ComSoc Asia-Pacific Best Young Researcher Award. He is an Editor of IEEE Transactions on Communications, IEEE Transactions on Machine Learning in Communications and Networking, and was an editor of IEEE Transactions on Wireless Communications (2015-2020). He served as a MAC track co-chair for IEEE Wireless Communications and Networking Conference (WCNC) 2011 and a co-chair for the Wireless Communications Symposium of IEEE International Conference on Communications (ICC) 2021. He is an IEEE Fellow and an IEEE ComSoc Distinguished Lecturer.
\end{IEEEbiography}
\vspace{-2cm}

\begin{IEEEbiography}
[{\includegraphics[width=1.4in,height=1.25in,clip,keepaspectratio]{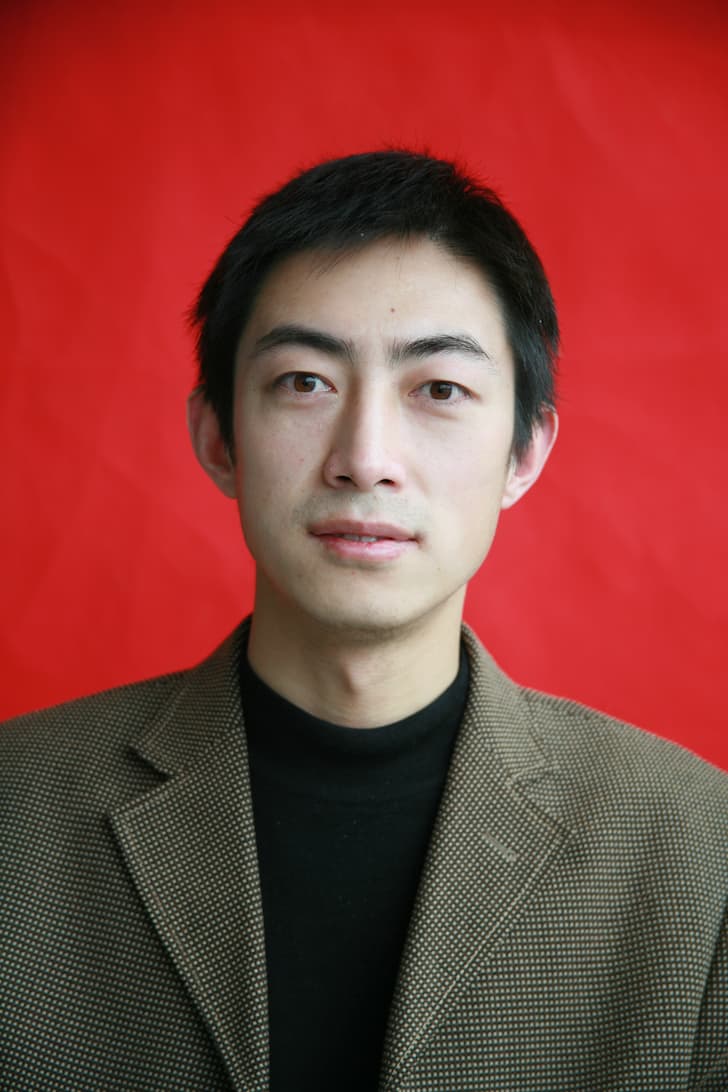}}]{Liang Wang} received the Ph.D. degree in computer science from the Shenyang Institute of Automation (SIA), Chinese Academy of Sciences, Shenyang, China, in 2014. He is currently an associate professor at Northwestern Polytechnical University, Xi'an, China. His research interests include ubiquitous computing, mobile crowd sensing, and data mining.
\end{IEEEbiography}
\vspace{-2cm}

\begin{IEEEbiography}[{\includegraphics[width=1.4in,height=1.25in,clip,keepaspectratio]{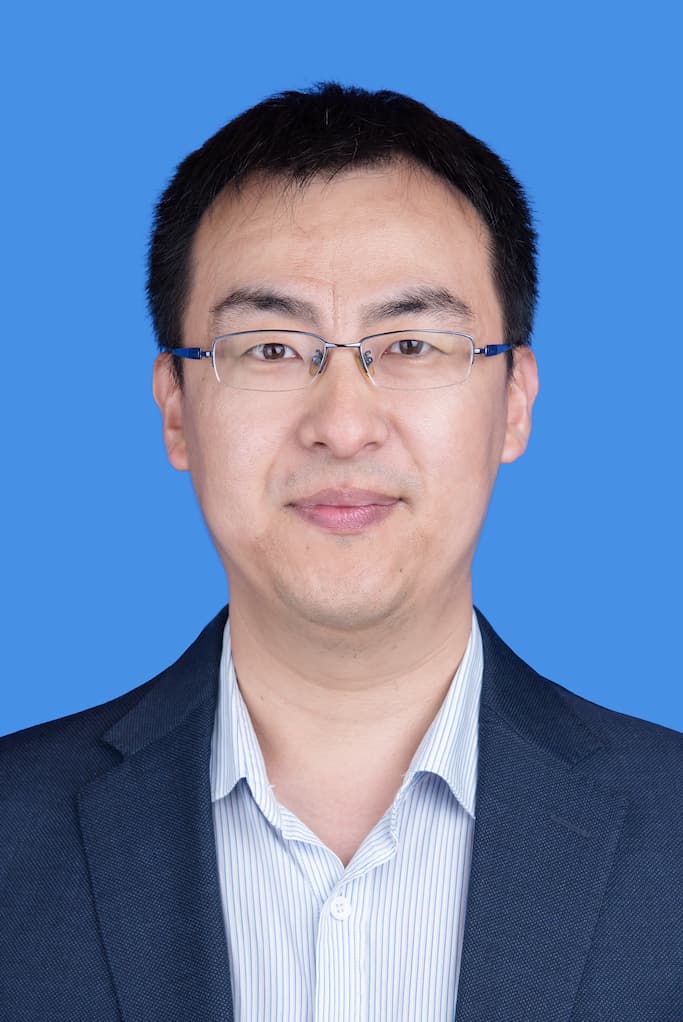}}]{Bin Guo} received the PhD degree in computer science from Keio University, Minato, Japan, in 2009, and then was a postdoc researcher with Insti-
tut Telecom SudParis, France. He is currently a professor with Northwestern Polytechnical University, Xi'an, China. His research interests include ubiquitous computing, mobile crowd sensing, and human-computer interaction. He has served as an associate editor of the IEEE Communications Magazine and the IEEE Transactions on Human-Machine-Systems, the guest editor of the ACM Transactions on Intelligent Systems and Technology and the IEEE Internet of Things, the general co-chair of IEEE UIC'15, and the program chair of IEEE CPSCom'16, ANT'14, and UIC'13. He is a senior member of the IEEE.
\end{IEEEbiography}

\end{document}